\pdfoutput=1 
\documentclass{article} 
\usepackage{iclr2022_conference,times}

\usepackage[utf8]{inputenc} 
\usepackage[T1]{fontenc}    
\usepackage[hidelinks]{hyperref}       
\usepackage{url}            
\usepackage{booktabs}       
\usepackage{amsfonts}       
\usepackage{nicefrac}       
\usepackage{microtype}      
\usepackage{xcolor}         
\usepackage{caption}

\iclrfinalcopy 

\usepackage{amsmath,mathtools}
\usepackage{amsthm}
\usepackage{booktabs}

\usepackage{tikz}
\usetikzlibrary{automata,arrows,shapes,calc,positioning,patterns}
\tikzset{every edge/.style={draw,->,>=stealth',shorten >=1pt,auto,semithick}}
\tikzset{initial text={},double distance=2pt}
\tikzset{state/.append style={minimum size=18pt,inner sep=1pt}}

\usepackage{pgfplots}
\DeclareUnicodeCharacter{2212}{\ensuremath{-}} 

\usepackage{enumitem}

\newcommand{\indicator}[1]{\mathbb{I}[#1]}

\newcommand{\transtensor}[5]{\ensuremath{\Delta[#1][#2,#3\rightarrow #4,#5]}}

\newcommand{\tprev}{t\mathord-1}

\newcommand{\reverse}[1]{{#1}^{\mathrm{R}}}

\newcommand{\Wparam}[1]{\mathbf{W}_{\mathrm{#1}}}
\newcommand{\bparam}[1]{\mathbf{b}_{\mathrm{#1}}}
\newcommand{\Affine}[2]{\Wparam{#1} #2 + \bparam{#1}}
\newcommand{\newmodel}{Renormalizing NS-RNN}
\newcommand{\newmodelacronym}{RNS-RNN}
\newcommand{\codeurl}[0]{https://github.com/bdusell/nondeterministic-stack-rnn}

\usepackage{newtxtext,newtxmath}

\title{Learning Hierarchical Structures with \\ Differentiable Nondeterministic Stacks}

\author{Brian DuSell and David Chiang \\
Department of Computer Science and Engineering,
University of Notre Dame \\
\texttt{\{bdusell1,dchiang\}@nd.edu}
}

\begin{document}

\maketitle

\begin{abstract}
Learning hierarchical structures in sequential data---from simple algorithmic patterns to natural language---in a reliable, generalizable way remains a challenging problem for neural language models. Past work has shown that recurrent neural networks (RNNs) struggle to generalize on held-out algorithmic or syntactic patterns without supervision or some inductive bias. To remedy this, many papers have explored augmenting RNNs with various differentiable stacks, by analogy with finite automata and pushdown automata (PDAs). In this paper, we improve the performance of our recently proposed Nondeterministic Stack RNN (NS-RNN), which uses a differentiable data structure that simulates a nondeterministic PDA, with two important changes. First, the model now assigns unnormalized positive weights instead of probabilities to stack actions, and we provide an analysis of why this improves training. Second, the model can directly observe the state of the underlying PDA. Our model achieves lower cross-entropy than all previous stack RNNs on five context-free language modeling tasks (within 0.05 nats of the information-theoretic lower bound), including a task on which the NS-RNN previously failed to outperform a deterministic stack RNN baseline. Finally, we propose a restricted version of the NS-RNN that incrementally processes infinitely long sequences, and we present language modeling results on the Penn Treebank.
\end{abstract}

\section{Introduction}

Many machine learning problems involve sequential data with hierarchical structures, such as modeling context-free languages \citep{grefenstette+al:2015,dusell+chiang:2020}, evaluating mathematical expressions \citep{nangia+bowman:2018,hao+al:2018}, logical inference \citep{bowman+al:2015}, and modeling syntax in natural language \citep{dyer+al:2016,shen+al:2018,kim+al:2019}. However, recurrent neural networks (RNNs) have difficulty learning to solve these tasks, or generalizing to held-out sequences, unless they have supervision or a hierarchical inductive bias \citep{schijndel+al:2019,wilcox+al:2019,mccoy+al:2020}. A limiting factor of RNNs is their reliance on memory whose size is constant across time. For example, to predict the second half of a string of the form~$w\texttt{\#}\reverse{w}$, a simple RNN would need to store all of $w$ in its hidden state before predicting its reversal $\reverse{w}$; a memory of finite size will inevitably fail to do this for inputs exceeding a certain length.

To remedy this, some previous work has investigated the addition of differentiable stack data structures to RNNs \citep{sun+al:1995,grefenstette+al:2015,joulin+mikolov:2015,dusell+chiang:2020}, which is closely related to work on neural networks that model shift-reduce parsers \citep{bowman+al:2016,dyer+al:2016,shen+al:2019}. Just as adding a stack to a finite state machine, which makes it a pushdown automaton (PDA), enables it to recognize context-free languages (CFLs), the hope is that adding stacks to RNNs will increase the range of problems on which they can be used effectively. We also expect stacks to aid training by introducing an inductive bias for learning hierarchical patterns, and to increase generalization power by structuring the model's memory in a way that better predicts held-out hierarchical data.

Previously \citep{dusell+chiang:2020}, we proposed a stack-based RNN called the Nondeterministic Stack RNN (NS-RNN) that outperformed other stack RNNs on a range of CFL language modeling tasks. Its defining feature is that its external data structure is a nondeterministic PDA, allowing it to simulate an exponential number of sequences of stack operations in parallel. This is in contrast to prior stack RNNs \citep{grefenstette+al:2015,joulin+mikolov:2015} which model deterministic stacks, being designed to learn \emph{one} correct stack operation at each time step. One reason nondeterminism is important is that deterministic CFLs are a proper subset of CFLs. If the analogy with PDAs holds true, then equipping an RNN with a deterministic stack would only enable it to model deterministic CFLs, whereas a nondeterministic stack should enable it to model all CFLs. This is important for natural language processing, as human language is known to be high in syntactic ambiguity.

Another benefit of nondeterminism, even on deterministic CFLs, applies to training.
In order for a model to receive a reward for an action, it must try the action (that is, give it nonzero probability so that it receives gradient during backpropagation). For example, in the digit-recognition task, a classifier tries all ten digits, and is rewarded for the correct one. But in a stack-augmented model, the space of possible action sequences is very large. Whereas a deterministic stack can only try one of them, a nondeterministic stack can try all of them and always receives a reward for the correct one.
But as explained in \S\ref{sec:why_unnormalized}, because the NS-RNN's probability for an action sequence is the product of many probabilities, it can be extremely small, so the NS-RNN sometimes learns very slowly.

In this paper we present a new model, the \newmodel{} (\newmodelacronym{}), which is based on the NS-RNN, but improves its performance on all of the CFL tasks it was originally tested on, thanks to two key changes. The first is that stack actions have weights that do not necessarily form a probability distribution (\S\ref{sec:why_unnormalized}). They define an unnormalized distribution over stacks that is \emph{renormalized} whenever the model queries it. Second, the \newmodelacronym{} includes not only top stack symbols but also PDA states in this query (\S\ref{sec:pda_states}). These changes allow the \newmodelacronym{} to attain lower cross-entropy on CFL tasks (in fact, very close to the information-theoretic lower bound) and to surpass deterministic stack RNNs on a task on which the NS-RNN fails to do so (``padded reversal''). Finally, as a third modification, we present a memory-restricted version of the RNS-RNN that requires only~$O(n)$ time and space~(\S\ref{sec:incremental_execution}). This restricted RNS-RNN can be run incrementally on arbitrarily long sequences, which is a necessity for language modeling on natural language, for which we provide experimental results. Our code is available at \url{\codeurl}.

\section{Previous stack RNNs}

We begin by discussing three previously proposed stack RNNs, each of which uses a different style of differentiable stack: stratification \citep{das+al:1992,sun+al:1995,grefenstette+al:2015}, superposition \citep{joulin+mikolov:2015}, and nondeterminism \citep{dusell+chiang:2020}.

\subsection{Controller-stack interface}

Each type of stack RNN consists of a simple RNN (or variant such as LSTM), called the \textit{controller}, connected to a differentiable stack. The stack has no parameters of its own; its role is to accept \textit{actions} from the controller to push and pop elements at each time step, simulate those actions, and return a \textit{reading} to the controller as an extra input to the next time step that serves as a representation of the updated top element of the stack. The stack actions and stack reading take on continuous values so that they may be differentiable; their form and interpretation vary with architecture.

Following prior work \citep{dusell+chiang:2020}, we make minor changes to the original model definitions given by \citet{grefenstette+al:2015} and \citet{joulin+mikolov:2015} to ensure that all three of these stack RNN models conform to the same controller-stack interface. This allows us to isolate differences in the style of stack data structure employed while keeping other parts of the network the same. We assume the input $w = w_1 \cdots w_n$ is encoded as a sequence of vectors $\mathbf{x}_1, \cdots, \mathbf{x}_n$. In all of our experiments, we use an LSTM \citep{hochreiter+al:1997} as the controller, whose memory consists of a hidden state $\mathbf{h}_t$ and memory cell $\mathbf{c}_t$. The controller computes the next state~$(\mathbf{h}_t, \mathbf{c}_t)$ given the previous state $(\mathbf{h}_{t-1}, \mathbf{c}_{t-1})$, input vector $\mathbf{x}_t$, and stack reading $\mathbf{r}_{t-1}$:
\[ (\mathbf{h}_t, \mathbf{c}_t) = \mathrm{LSTM}\left((\mathbf{h}_{t-1}, \mathbf{c}_{t-1}), \begin{bmatrix}
\mathbf{x}_t \\
\mathbf{r}_{t-1}
\end{bmatrix}\right). \]
We set $\mathbf{h}_0 = \mathbf{c}_0 = \mathbf{0}$. The hidden state is used to compute the stack actions $a_t$ and predict the logits $\mathbf{y}_t$ for the next word $w_{t+1}$. The previous stack and new actions are used to compute a new stack $s_t$, which in turn is used to produce a new reading $\mathbf{r}_t$:
\begin{align*}
& a_t = \textsc{Actions}(\mathbf{h}_t) && \mathbf{y}_t = \Affine{hy}{\mathbf{h}_t} && s_t = \textsc{Stack}(s_{t-1}, a_t) && \mathbf{r}_t = \textsc{Reading}(s_t)
\end{align*}
In order to change the stack data structure, we need only change the definitions of \textsc{Actions}, \textsc{Stack}, \textsc{Reading}, and $s_0$, which may depend on parameters of the model; for our changes to the NS-RNN, we will only need to change \textsc{Actions} and \textsc{Reading}.

\subsection{Stratification}

Based on work by \citet{das+al:1992} and \citet{sun+al:1995}, the stack of \citet{grefenstette+al:2015} relies on a strategy we have dubbed ``stratification'' \citep{dusell+chiang:2020}. The elements of the stack are vectors, each of which is associated with a ``thickness'' between 0 and 1, which represents the degree to which the vector element is present on the stack. A helpful analogy is that of layers of a cake; the stack elements are like cake layers of varying thickness.
In this model, $a_t = (u_t, d_t, \mathbf{v}_t)$, where the pop signal $u_t \in (0, 1)$ indicates the amount to be removed from the top of the stack, $\textbf{v}_t$ is a learned vector to be pushed as a new element onto the stack, and the push signal $d_t \in (0, 1)$ is the thickness of that newly pushed vector. This model has quadratic time and space complexity with respect to input length. We refer the reader to Appendix \ref{sec:stratification_details} for full details.

\subsection{Superposition}

The stack of \citet{joulin+mikolov:2015} simulates a combination of partial stack actions by computing three new, separate stacks: one with all cells shifted down (push), kept the same (no-op), and shifted up (pop). The new stack is then an element-wise interpolation (``superposition'') of these three stacks. In this model, stack elements are again vectors, and $a_t = (\mathbf{a}_t, \mathbf{v}_t)$, where the vector $\mathbf{a}_t$ is a probability distribution over three stack operations: push a new vector, no-op, and pop the top vector; $\mathbf{v}_t$ is the vector to be pushed. The vector $\mathbf{v}_t$ can be learned or can be set to $\mathbf{h}_t$ \citep{yogatama+al:2018}. The stack reading is the top cell of the stack. This model has quadratic time and space complexity with respect to input length. We refer the reader to Appendix \ref{sec:superposition_details} for full details.

\subsection{Nondeterminism}
\label{sec:nondeterminism}

The stack module in the Nondeterministic Stack RNN (NS-RNN) model \citep{dusell+chiang:2020} maintains a probability distribution over whole stacks by simulating a weighted PDA. It has cubic time complexity and quadratic space complexity with respect to input length, leading to higher wall-clock run time than other stack RNNs, but often better task performance.

The simulated weighted PDA maintains a state drawn from a finite set $Q$, which includes an initial state $q_0$, and a stack with symbols drawn from an alphabet $\Gamma$, which includes an initial symbol $\bot$. At each time step, the PDA executes a weighted \textit{transition} that changes its state and manipulates the stack. Stack operations are drawn from the set $\text{Op}(\Gamma) = \mathord\bullet\Gamma \cup \Gamma \cup \{\epsilon\}$, where for any $y \in \Gamma$,  $\bullet y$ means ``push~$y$,'' $y$ means ``replace top element with $y$,'' and $\epsilon$ means ``pop top element.'' A valid sequence of transitions is called a \textit{run}, and the weight of a run is the product of the weights of its transitions.

The RNN controller emits transition weights to the stack module. Note that the stack module, not the controller, keeps track of PDA states and stack configurations, so the controller emits \emph{distributions} over transitions conditioned on the PDA's current state and top stack symbol. More precisely, $a_t=\Delta[t]$ is a tensor where the meaning of element $\transtensor tqxr\upsilon$ is: if the PDA is in state $q$ and the top stack symbol is $x$, then, with weight $\transtensor tqxr\upsilon$, go to state $r$ and perform stack operation $\upsilon \in \text{Op}(\Gamma)$. The original NS-RNN definition requires that for all $t$, $q$, and $x$, the weights form a probability distribution. Accordingly, they are computed from the hidden state using a softmax layer:
\begin{equation} \label{eq:normalized}
    \Delta[t] = \underset{q, x}{\mathrm{softmax}}(\Affine{ha}{\mathbf{h}_t}).
\end{equation}

The stack module marginalizes over all runs ending at time step $t$ and returns the distribution over top stack symbols at $t$ to the controller. It may appear that computing this distribution is intractable because the number of possible runs is exponential in $t$, but \citet{lang:1974} gives a dynamic programming algorithm that simulates all runs of a nondeterministic PDA in cubic time and quadratic space. Lang's algorithm exploits structural similarities in PDA runs. First, multiple runs can result in the same stack. Second, for $k > 0$, a stack of height $k$ must have been derived from a stack of height~$k-1$, so in principle representing a stack of height $k$ requires only storing its top symbol and a pointer to a stack of height $k-1$. The resulting data structure is a weighted graph where edges represent individual stack symbols, and paths (of which there are exponentially many) represent stacks.

We may equivalently view this graph as a weighted finite automaton (WFA) that encodes a distribution over stacks, and accordingly the NS-RNN's stack module is called the \textit{stack WFA}. Indeed, the language of stacks at a given time step $t$ is always regular \citep{autebert+:1997}, and Lang's algorithm gives an explicit construction for the WFA encoding this language. Its states are PDA configurations of the form $(i, q, x)$, where $0 \leq i \leq n$, $q \in Q$, and $x \in \Gamma$ is the stack top. A stack WFA transition from $(i, q, x)$ to $(t, r, y)$ means the PDA went from configuration $(i, q, x)$ to $(t, r, y)$ (possibly via multiple time steps), where the only difference in the stack is that a single $y$ was pushed, and the $x$ was never modified in between. The weights of these transitions are stored in a tensor $\gamma$ of shape~$n \times n \times |Q| \times |\Gamma| \times |Q| \times |\Gamma|$, where elements are written $\gamma[i \xrightarrow{} t][q, x \xrightarrow{} r, y]$. For $0 \leq i < t \leq n$,
\begin{equation} \label{eq:gamma}
\begin{split}
    &\gamma[i \xrightarrow{} t][q, x \xrightarrow{} r, y] = \\
&\qquad \begin{aligned}
    &\indicator{i=\tprev} \; \transtensor tqxr{\bullet y} && \text{push} \\
                      & + \sum_{s,z} \gamma[i \xrightarrow{} \tprev][q, x \xrightarrow{} s, z] \; \transtensor tszry && \text{repl.} \\
                      & + \sum_{k=i+1}^{t-2} \sum_{u} \sum_{s,z}  \gamma[i \xrightarrow{} k][q, x \xrightarrow{} u, y] \; \gamma[k \xrightarrow{} \tprev][u, y \xrightarrow{} s, z] \; \transtensor tszr\epsilon && \text{pop}
\end{aligned}
\end{split}
\end{equation}

The NS-RNN sums over all stacks (accepting paths in the stack WFA) using a tensor $\alpha$ of \textit{forward weights} of shape $n \times |Q| \times |\Gamma|$. The weight $\alpha[t][r, y]$ is the total weight of reaching configuration~$(t, r, y)$. These weights are normalized to get the distribution over top stack symbols at~$t$:
\begin{align} 
    \alpha[0][r, y] &= \indicator{r = q_0 \wedge y = \bot} \label{eq:alpha1} \\
    \alpha[t][r, y] &= 
    \sum_{i=0}^{t-1} \sum_{q,x} \alpha[i][q, x] \, \gamma[i \xrightarrow{} t][q, x \xrightarrow{} r, y] \qquad (1 \leq t \leq n)
\label{eq:alpha2} \\
    \mathbf{r}_t[y] &= \frac{ \sum_r \alpha[t][r, y] }{ \sum_{y'} \sum_r \alpha[t][r, y'] }.
\label{eq:original_reading}
\end{align}

We refer the reader to our earlier paper \citep{dusell+chiang:2020} for details of deriving these equations from Lang's algorithm. To avoid underflow and overflow, in practice, $\Delta$, $\gamma$, and $\alpha$ are computed in log space. The model's time complexity is $O({|Q|}^4 {|\Gamma|}^3 n^3)$, and its space complexity is $O({|Q|}^2 {|\Gamma|}^2 n^2)$.

\section{\newmodel{}}

Here, we introduce the \newmodel{}, which differs from the NS-RNN in two ways.

\subsection{Unnormalized transition weights}

\label{sec:why_unnormalized}

To make a good prediction at time $t$, the model may need a certain top stack symbol $y$, which may in turn require previous actions to be orchestrated correctly. For example, consider the language $\{v\texttt{\#}\reverse{v}\}$, where $n$ is odd and $w_t = w_{n-t+1}$ for all $t$. In order to do better than chance when predicting $w_t$ (for $t$ in the second half), the model has to push a stack symbol that encodes $w_t$ at time $(n-t+1)$, and that same symbol must be on top at time $t$. How does the model learn to do this? Assume that the gradient of the log-likelihood with respect to $\mathbf{r}_t[y]$ is positive; this gradient ``flows'' to the PDA transition probabilities via (among other things) the partial derivatives of $\log \alpha$ with respect to $\log \Delta$.

To calculate these derivatives more easily, we express $\alpha$ directly (albeit less efficiently) in terms of $\Delta$:
\begin{equation*}
\alpha[t][r,y] = \sum_{\delta_1 \cdots \delta_t \leadsto r, y} \prod_{i=1, \ldots, t} \Delta[i][\delta_i]
\end{equation*}
where each $\delta_i$ is a PDA transition of the form $q_1, x_1 \rightarrow q_2, x_2$, $\Delta[i][\delta_i] = \Delta[i][q_1, x_1 \rightarrow q_2, x_2]$, and the summation over $\delta_1 \cdots \delta_t \leadsto r, y$ means that after following transitions $\delta_1, \ldots, \delta_t$, then the PDA will be in state $r$ and its top stack symbol will be $y$. Then the partial derivatives are:
\begin{align*}
\frac{\partial \log \alpha[t][r,y]}{\partial \log \Delta[i][\delta]} &=  \frac{\sum _{\delta_1 \cdots \delta_t \leadsto r,y} 
\left(\prod_{i'=1}^{t} \Delta[i'][\delta_{i'}]\right) \, \indicator{\delta_i=\delta}}{\sum _{\delta_1 \cdots \delta_t \leadsto r,y} 
\prod_{i'=1}^{t} \Delta[i][\delta_{i'}]}.
\end{align*}
This is the posterior probability of having used transition $\delta$ at time $i$, given that the PDA has read the input up to time $t$ and reached state $r$ and top stack symbol $y$. 

So if a correct prediction at time $t$ depends on a stack action at an earlier time $i$, the gradient flow to that action is proportional to its probability given the correct prediction. This probability is always nonzero, as desired. However, this probability is the product of individual action probabilities, which are always strictly less than one. If a correct prediction depends on orchestrating many stack actions, then this probability may become very small. Returning to our example, we expect the model to begin by learning to predict the middle of the string, where only a few stack actions must be orchestrated, then working its way outwards, more and more slowly as more and more actions must be orchestrated.
In \S\ref{sec:cfl_experiments} we verify empirically that this is the case. 

The solution we propose is to use unnormalized (non-negative) transition weights, not probabilities, and to normalize weights only when reading.
Equation (\ref{eq:normalized}) now becomes
\begin{equation*}
    \Delta[t] = \exp (\Affine{ha}{\mathbf{h}_t}).
\end{equation*}
The gradient flowing to a transition is still proportional to its posterior probability, but now each transition weight has the ability to ``amplify'' \citep{lafferty+:2001} other transitions in shared runs. Equation (\ref{eq:original_reading}) is not changed (yet), but its interpretation is. The NS-RNN maintains a probability distribution over stacks and updates it by performing probabilistic operations. Now, the model maintains an unnormalized weight distribution, and when it reads from the stack at each time step, it renormalizes this distribution and marginalizes it to get a probability distribution over readings. For this reason, we call our new model a \newmodel{} (\newmodelacronym{}).

\subsection{PDA states included in stack reading}
\label{sec:pda_states}

In the NS-RNN, the controller can read the distribution over the PDA's current top stack symbol, but it cannot observe its current state. To see why this is a problem, consider the language $\{v\reverse{v}\}$. While reading $v$, the controller should predict the uniform distribution, but while reading $\reverse{v}$, it should predict based on the top stack symbol. A PDA with two states can nondeterministically guess whether the current position is in $v$ or $\reverse{v}$. The controller should interpolate the two distributions based on the weight of being in each state, but it cannot do this without input from the stack WFA, since the state is entangled with the stack contents. We solve this in the \newmodelacronym{} by computing a joint distribution over top stack symbols \emph{and} PDA states, making $\mathbf{r}_t$ a vector of size $|Q| |\Gamma|$. Equation~\ref{eq:original_reading} becomes
\begin{equation*}
    \mathbf{r}_t[(r, y)] = \frac{ \alpha[t][r, y] }{ \sum_{r', y'} \alpha[t][r', y'] }.
\end{equation*}

\section{Experiments on formal languages}
\label{sec:cfl_experiments}

In order to assess the benefits of using unnormalized transition weights and including PDA states in the stack reading, we ran the \newmodelacronym{} with and without the two proposed modifications on the same five CFL language modeling tasks used previously \citep{dusell+chiang:2020}. We use the same experimental setup and PCFG settings, except for one important difference: we require the model to predict an end-of-sequence (EOS) symbol at the end of every string. This way, the model defines a proper probability distribution over strings, improving the interpretability of the results.

Each task is a weighted CFL specified as a PCFG:
\begin{description}
    \item[Marked reversal] The palindrome language with a middle marker ($\{v\texttt{\#}\reverse{v} \mid v \in \{\texttt{0}, \texttt{1}\}^{*}\}$).
    \item[Unmarked reversal] The palindrome language without a middle marker ($\{v\reverse{v} \mid v \in \{\texttt{0}, \texttt{1}\}^{*}\}$).
    \item[Padded reversal] Like unmarked reversal, but with a long stretch of repeated symbols in the middle ($\{va^p\reverse{v} \mid v \in \{\texttt{0}, \texttt{1}\}^{*} , a \in \{\texttt{0}, \texttt{1}\}, p \geq 0\}$).
    \item[Dyck language] The language $D_2$ of strings with balanced brackets (two bracket types).
    \item[Hardest CFL] A language shown by \citet{greibach:1973} to be at least as hard to parse as any other CFL.
\end{description}
The marked reversal and Dyck languages are deterministic tasks that could be solved optimally with a deterministic PDA. On the other hand, the unmarked reversal, padded reversal, and hardest CFL tasks require nondeterminism, with hardest CFL requiring the most \citep[Appendix A]{dusell+chiang:2020}. We randomly sample from these languages to create training, validation, and test sets.
All strings are represented as sequences of one-hot vectors.
Please see Appendix~\ref{sec:formal_hyperparams} for additional experimental details.

We evaluate models according to per-symbol cross-entropy (lower is better). For any set of strings $S$ and probability distribution $p$, it is defined as
\begin{equation*}
    H(S, p) = \frac{-\sum_{w \in S}  \log p(w \cdot \mathrm{EOS})}{\sum_{w \in S} (|w| + 1)}.
\end{equation*}
Since the validation and test strings are all sampled from known distributions, we can also use this formula to compute the per-symbol entropy of the true distribution  \citep{dusell+chiang:2020}. In our experiments we measure performance as the difference between the model cross-entropy and the true entropy, per-symbol and measured in nats (lower is better, and zero is optimal).

We compare seven models on the CFL tasks, each of which consists of an LSTM connected to a different type of stack: none (``LSTM''); stratification (``Gref''); superposition (``JM''); nondeterministic, aka NS-RNN (``NS''); NS with PDA states in the reading and normalized action weights (``NS+S''); NS with no states in the reading and unnormalized action weights (``NS+U''); and NS with PDA states and unnormalized action weights, or \newmodelacronym{} (``NS+S+U'').

\paragraph{Results} \label{sec:cfl_results} We show validation set performance as a function of training time in Figure \ref{fig:cfl-train}, and test performance binned by string length in Figure \ref{fig:cfl-test} (see also Appendix~\ref{sec:wall_clock_time} for wall-clock training times). For all tasks, we see that our \newmodelacronym{} (denoted NS+S+U) attains near-optimal cross-entropy (within 0.05 nats) on the validation set. All stack models effectively solve the deterministic marked reversal and Dyck tasks, although we note that on marked reversal the NS models do not generalize well on held-out lengths. Our new model excels on the three nondeterministic tasks: unmarked reversal, padded reversal, and hardest CFL. We find that the combination of both enhancements (+S+U) greatly improves performance on unmarked reversal and hardest CFL over previous work. For unmarked reversal, merely changing the task by adding EOS causes the baseline NS model to perform worse than Gref and JM; this may be because it requires the NS-RNN to learn a correlation between the two most distant time steps. Both enhancements (+S+U) in the \newmodelacronym{} are essential here; without unnormalized weights, the model does not find a good solution during training, and without PDA states, the model does not have enough information to make optimal decisions. For padded reversal, we see that the addition of PDA states in the stack reading (+S) proves essential to improving performance. Although NS+S and NS+S+U have comparable performance on padded reversal, NS+S+U converges much faster. On hardest CFL, using unnormalized weights by itself (+U) improves performance, but only both modifications together (+S+U) achieve the best performance.

\begin{figure*}
\pgfplotsset{
  every axis/.style={
    height=2.1in,width=3.4in,
    ytick distance=0.1
  },
  title style={yshift=-4.5ex},
}
    \centering
\begin{tabular}{@{}l@{\hspace{1em}}l@{}}    
    \multicolumn{2}{c}{
    \begin{tikzpicture}

\definecolor{color0}{rgb}{0.12156862745098,0.466666666666667,0.705882352941177}
\definecolor{color1}{rgb}{1,0.498039215686275,0.0549019607843137}
\definecolor{color2}{rgb}{0.172549019607843,0.627450980392157,0.172549019607843}
\definecolor{color3}{rgb}{0.83921568627451,0.152941176470588,0.156862745098039}
\definecolor{color4}{rgb}{0.580392156862745,0.403921568627451,0.741176470588235}
\definecolor{color5}{rgb}{0.549019607843137,0.337254901960784,0.294117647058824}
\definecolor{color6}{rgb}{0.890196078431372,0.466666666666667,0.76078431372549}

\begin{axis}[
  hide axis,
  height=2cm,
  legend style={
    draw=none,
    /tikz/every even column/.append style={column sep=0.4cm}},
  legend columns=-1,
  xmin=0,xmax=1,ymin=0,ymax=1
    ]
\addlegendimage{color0}
\addlegendentry{LSTM}
\addlegendimage{color1}
\addlegendentry{Gref}
\addlegendimage{color2}
\addlegendentry{JM}
\addlegendimage{color3}
\addlegendentry{NS}
\addlegendimage{color4}
\addlegendentry{NS+S}
\addlegendimage{color5}
\addlegendentry{NS+U}
\addlegendimage{color6}
\addlegendentry{NS+S+U}
\end{axis}

\end{tikzpicture}
    } \\
    \scalebox{0.8}{\input{figures/cfl/train/marked-reversal}}
    &\scalebox{0.8}{
\begin{tikzpicture}

\definecolor{color0}{rgb}{0.12156862745098,0.466666666666667,0.705882352941177}
\definecolor{color1}{rgb}{1,0.498039215686275,0.0549019607843137}
\definecolor{color2}{rgb}{0.172549019607843,0.627450980392157,0.172549019607843}
\definecolor{color3}{rgb}{0.83921568627451,0.152941176470588,0.156862745098039}
\definecolor{color4}{rgb}{0.580392156862745,0.403921568627451,0.741176470588235}
\definecolor{color5}{rgb}{0.549019607843137,0.337254901960784,0.294117647058824}
\definecolor{color6}{rgb}{0.890196078431372,0.466666666666667,0.76078431372549}

\begin{axis}[
legend cell align={left},
legend style={
  fill opacity=0.8,
  draw opacity=1,
  text opacity=1,
  at={(0.03,0.97)},
  anchor=north west,
  draw=white!80!black
},
tick align=outside,
tick pos=left,
title={Marked Reversal},
x grid style={white!69.0196078431373!black},
xmin=38.1, xmax=101.9,
xtick style={color=black},
y grid style={white!69.0196078431373!black},
ymin=0, ymax=0.289548975949523,
ytick style={color=black},
]
\addplot [semithick, color0]
table {%
41 0.0922771517422872
43 0.0762958435999411
45 0.0683919691819287
47 0.062378898182213
49 0.0627103494249756
51 0.0631964050793534
53 0.0669109332460219
55 0.0727368905531281
57 0.0783727023670733
59 0.0859701957371723
61 0.0871942510899461
63 0.0994670623825842
65 0.107789353802848
67 0.114590317304089
69 0.129449640926687
71 0.140698310665303
73 0.141296648488151
75 0.14672658335404
77 0.159592277770074
79 0.169392354701637
81 0.175680753194843
83 0.188013204187259
85 0.19891318879994
87 0.205711096697271
89 0.215773118844305
91 0.217748948282499
93 0.240523138369601
95 0.25275337213146
97 0.270814231715536
99 0.277440401056877
};
\addlegendentry{LSTM}
\addplot [semithick, color1]
table {%
41 0.0904587691157991
43 0.0719427607696854
45 0.0639283089832194
47 0.0597993744598823
49 0.0570854714952881
51 0.0554006193221419
53 0.0527349530666237
55 0.0512470659856057
57 0.0495940789834526
59 0.0480043347345681
61 0.0460815998802687
63 0.0442768615769201
65 0.0422692232985864
67 0.0410022758863495
69 0.0403842809936518
71 0.0413898389856158
73 0.0430851765181344
75 0.0466941704736943
77 0.0509344878061317
79 0.0559499413715585
81 0.0616852132272365
83 0.0670425417523634
85 0.0727524136818438
87 0.0782318763936486
89 0.0839765964918742
91 0.0895394013225809
93 0.0955806412785102
95 0.100784530578725
97 0.106427473105842
99 0.111870869806877
};
\addlegendentry{Gref}
\addplot [semithick, color2]
table {%
41 0.0769825553881206
43 0.0694857905548417
45 0.0660382677977574
47 0.0634110281057156
49 0.0605053689562257
51 0.0578273771346418
53 0.0559864540793553
55 0.0533631984493335
57 0.0513322760470301
59 0.0494319063491514
61 0.0473301035345025
63 0.0453473800705726
65 0.0434897414685674
67 0.0414783860081327
69 0.0397181956141874
71 0.0384746981978554
73 0.0383499371283541
75 0.0401707008499277
77 0.0448857473213562
79 0.0529983422504648
81 0.0630402234692334
83 0.0725058567170212
85 0.0826240979114949
87 0.0921034195843446
89 0.101605801135972
91 0.1111107116147
93 0.120296710926117
95 0.137500839172475
97 0.149525617637092
99 0.168520137385002
};
\addlegendentry{JM}
\addplot [semithick, color3]
table {%
41 0.107618512651885
43 0.0887366394983787
45 0.0752517369298906
47 0.0699329397268093
49 0.0634162820421631
51 0.0599081595114447
53 0.0551596763739154
55 0.0534553179387308
57 0.0494897720060819
59 0.0470205293960264
61 0.0439172144846034
63 0.0412669509934241
65 0.0387409473752909
67 0.0372638725660371
69 0.0374573836722232
71 0.0375790072798867
73 0.0410760311422899
75 0.0442213793038752
77 0.0497009203481991
79 0.0552894799457773
81 0.0621117447630598
83 0.0684805009692534
85 0.0769464373464514
87 0.0845452697263901
89 0.0926066149380549
91 0.100940529676861
93 0.110376446215345
95 0.118670221903595
97 0.127579517908138
99 0.135686690119377
};
\addlegendentry{NS}
\addplot [semithick, color4]
table {%
41 0.0773815799881949
43 0.0652966148623701
45 0.0615380395793471
47 0.0590150981333848
49 0.0567897683702882
51 0.0549220332719615
53 0.053133444820096
55 0.0517645569175255
57 0.0504919692717068
59 0.0494234834975889
61 0.0486812643837969
63 0.0481133026169592
65 0.0475599355973554
67 0.0466053750361657
69 0.0446660541521339
71 0.0422686435103555
73 0.0397411097951614
75 0.0384571870238588
77 0.0404032628461959
79 0.0490165307270273
81 0.0629594784235017
83 0.0803801184822742
85 0.0993532380822799
87 0.120572824325112
89 0.143654900528333
91 0.167776865232228
93 0.192782893605638
95 0.218397089579376
97 0.245088536163878
99 0.270952998713127
};
\addlegendentry{NS+S}
\addplot [semithick, color5]
table {%
41 0.100022222426811
43 0.0788609570143586
45 0.0666519364883145
47 0.0600626394338406
49 0.0561915994249757
51 0.0534067181427548
53 0.0503903892645405
55 0.0484191764069228
57 0.048081501531944
59 0.0472564506199847
61 0.0463828772870227
63 0.0448978179978187
65 0.0435606902138327
67 0.042752979585798
69 0.0416785750784731
71 0.0409190544804074
73 0.0415673146621378
75 0.0432120184383324
77 0.046687473833776
79 0.0531655785785898
81 0.0622184520801328
83 0.0738540070611433
85 0.0879078688053904
87 0.104695749129799
89 0.120395406170694
91 0.137634042966603
93 0.154797573210026
95 0.169975611307892
97 0.186611126645383
99 0.202217110041252
};
\addlegendentry{NS+U}
\addplot [semithick, color6]
table {%
41 0.0744238163163199
43 0.0648460644362336
45 0.0622683384923906
47 0.0605987315562364
49 0.0599094216906007
51 0.0587995028482837
53 0.0580379587089849
55 0.0569872044482174
57 0.0559816395976767
59 0.0539487113621722
61 0.0517818109437768
63 0.0486489624069983
65 0.0459376581352341
67 0.0436386356060187
69 0.0415188024780268
71 0.0392349927942097
73 0.0366454726541142
75 0.0352689032039574
77 0.0386708660112601
79 0.0490094811664804
81 0.0631918586159561
83 0.0781729419104736
85 0.0943611016814805
87 0.10870825578818
89 0.122354336292221
91 0.136487271991738
93 0.151617200038484
95 0.163076858459585
97 0.175496251007628
99 0.187324385431877
};
\addlegendentry{NS+S+U}
\legend{}
\end{axis}

\end{tikzpicture}}
    \\ 
    \scalebox{0.8}{\input{figures/cfl/train/unmarked-reversal}}
    &\scalebox{0.8}{
\begin{tikzpicture}

\definecolor{color0}{rgb}{0.12156862745098,0.466666666666667,0.705882352941177}
\definecolor{color1}{rgb}{1,0.498039215686275,0.0549019607843137}
\definecolor{color2}{rgb}{0.172549019607843,0.627450980392157,0.172549019607843}
\definecolor{color3}{rgb}{0.83921568627451,0.152941176470588,0.156862745098039}
\definecolor{color4}{rgb}{0.580392156862745,0.403921568627451,0.741176470588235}
\definecolor{color5}{rgb}{0.549019607843137,0.337254901960784,0.294117647058824}
\definecolor{color6}{rgb}{0.890196078431372,0.466666666666667,0.76078431372549}

\begin{axis}[
legend cell align={left},
legend style={
  fill opacity=0.8,
  draw opacity=1,
  text opacity=1,
  at={(0.03,0.97)},
  anchor=north west,
  draw=white!80!black
},
tick align=outside,
tick pos=left,
title={Unmarked Reversal},
x grid style={white!69.0196078431373!black},
xmin=37, xmax=103,
xtick style={color=black},
y grid style={white!69.0196078431373!black},
ymin=0, ymax=0.472130997068097,
ytick style={color=black},
]
\addplot [semithick, color0]
table {%
40 0.271155934216185
42 0.233890534821015
44 0.219537930042221
46 0.210450467575053
48 0.208192609824592
50 0.208556691897698
52 0.213393544825546
54 0.225904017256367
56 0.226386659453557
58 0.228104870002845
60 0.233948812327076
62 0.24139374690453
64 0.246400213152219
66 0.25257860562206
68 0.252428070675172
70 0.263028376000378
72 0.274459042377972
74 0.266582091444593
76 0.287959706144118
78 0.295257913829585
80 0.302939772427958
82 0.30649906781075
84 0.306719221719645
86 0.331130661206722
88 0.320913422058145
90 0.326778770172966
92 0.329445829205564
94 0.335399902282185
96 0.348950998521238
98 0.34153529502273
100 0.365558229766116
};
\addlegendentry{LSTM}
\addplot [semithick, color1]
table {%
40 0.452125946601855
42 0.423011231132789
44 0.383319831083888
46 0.315105190397526
48 0.210026006269108
50 0.169384903064978
52 0.139060002667527
54 0.12378705170239
56 0.117054919459039
58 0.140642856628904
60 0.135990428297875
62 0.133862512405522
64 0.148792040075295
66 0.15295020951665
68 0.161027835734049
70 0.181230950824322
72 0.181267288311191
74 0.184501036757093
76 0.20208251204915
78 0.231018272809015
80 0.214652795672557
82 0.246076027995238
84 0.247875012160821
86 0.253199402226837
88 0.269341650200701
90 0.269800533565823
92 0.297761468906505
94 0.325972424897317
96 0.322670443189408
98 0.344832426493689
100 0.360534637563146
};
\addlegendentry{Gref}
\addplot [semithick, color2]
table {%
40 0.393475270153685
42 0.374773529007062
44 0.367160054608194
46 0.307538856870265
48 0.302742892828419
50 0.22997280326412
52 0.187398464489461
54 0.157692635864322
56 0.133866999708491
58 0.122051299756552
60 0.103264978438757
62 0.114655969436772
64 0.132006696024815
66 0.126253796561892
68 0.148760229302889
70 0.161989265359973
72 0.167241342517098
74 0.165020991184177
76 0.183558104327965
78 0.202306037346199
80 0.219198272138607
82 0.229289506910901
84 0.229848587528468
86 0.234352307219653
88 0.24537768425969
90 0.252434193850851
92 0.261675121880295
94 0.267866801953238
96 0.282924903160413
98 0.274936001304295
100 0.295078747247552
};
\addlegendentry{JM}
\addplot [semithick, color3]
table {%
40 0.218366359616337
42 0.195390523465637
44 0.183778435684583
46 0.187228486607633
48 0.187659835191302
50 0.199854610479194
52 0.191476396019532
54 0.203847332242162
56 0.199757984494675
58 0.214692156483247
60 0.208223674647773
62 0.213719787021098
64 0.222922486289238
66 0.22642334676012
68 0.226813119361766
70 0.244863350692279
72 0.250622556397664
74 0.24942278154876
76 0.255770430575936
78 0.273224995020249
80 0.285070487127032
82 0.281624491380027
84 0.296307514458616
86 0.300858963559452
88 0.306233657749859
90 0.312423569709367
92 0.315824724954892
94 0.325960963137449
96 0.333006300148171
98 0.355040266613639
100 0.36845905549322
};
\addlegendentry{NS}
\addplot [semithick, color4]
table {%
40 0.309690673640727
42 0.220217524283225
44 0.188433275094305
46 0.182383957435425
48 0.181349497579694
50 0.183676895252478
52 0.182217477977079
54 0.189395006247844
56 0.197700478669566
58 0.209250055053162
60 0.206368085828613
62 0.204707665245304
64 0.217648260027219
66 0.220933753709747
68 0.229455923936042
70 0.233523843925114
72 0.246227623820609
74 0.247482286757093
76 0.252968774224475
78 0.267250101329585
80 0.270886157941076
82 0.275056637934997
84 0.291615476315233
86 0.296940253968935
88 0.304007534154353
90 0.309604255234779
92 0.332408226298978
94 0.333167634601264
96 0.328757850566985
98 0.336832850657199
100 0.356685095870696
};
\addlegendentry{NS+S}
\addplot [semithick, color5]
table {%
40 0.261599496049264
42 0.187689079061568
44 0.168890631865138
46 0.171738698958032
48 0.18406668308671
50 0.185270176119512
52 0.17851340405314
54 0.190922616333071
56 0.196652130259478
58 0.194945442036743
60 0.198948051889064
62 0.201392041795403
64 0.210933979678661
66 0.219198606205082
68 0.230551619984502
70 0.226595786321681
72 0.242524177758965
74 0.241692443007093
76 0.253293322722851
78 0.268766245514711
80 0.26767393041407
82 0.269304155348401
84 0.278424242399792
86 0.311220684554423
88 0.311190315931039
90 0.312944528243021
92 0.30934717539172
94 0.32731139488087
96 0.338311803128197
98 0.332889116440028
100 0.352997702228987
};
\addlegendentry{NS+U}
\addplot [semithick, color6]
table {%
40 0.227940244905971
42 0.163629588236713
44 0.130952372316527
46 0.0824921325102256
48 0.0638520884597966
50 0.0638488468456153
52 0.0637615062052628
54 0.0572665882790943
56 0.0520249372770218
58 0.0604469234323998
60 0.0604414720991049
62 0.0583716347393515
64 0.0634914466257762
66 0.0725904482643175
68 0.0734049516194655
70 0.0874076398648148
72 0.0981338938869616
74 0.107282221652927
76 0.120410560953371
78 0.136101805987417
80 0.146089729025181
82 0.155976548044184
84 0.163603470340968
86 0.176720302909309
88 0.188879977535673
90 0.197810492357032
92 0.215221408842661
94 0.241352050719685
96 0.240119098150748
98 0.2385509594798
100 0.26173189351921
};
\addlegendentry{NS+S+U}
\legend{}
\end{axis}

\end{tikzpicture}}
    \\ 
    \scalebox{0.8}{\input{figures/cfl/train/padded-reversal}}
    &\scalebox{0.8}{\input{figures/cfl/test/padded-reversal}}
    \\ 
    \scalebox{0.8}{
\begin{tikzpicture}

\definecolor{color0}{rgb}{0.12156862745098,0.466666666666667,0.705882352941177}
\definecolor{color1}{rgb}{1,0.498039215686275,0.0549019607843137}
\definecolor{color2}{rgb}{0.172549019607843,0.627450980392157,0.172549019607843}
\definecolor{color3}{rgb}{0.83921568627451,0.152941176470588,0.156862745098039}
\definecolor{color4}{rgb}{0.580392156862745,0.403921568627451,0.741176470588235}
\definecolor{color5}{rgb}{0.549019607843137,0.337254901960784,0.294117647058824}
\definecolor{color6}{rgb}{0.890196078431372,0.466666666666667,0.76078431372549}

\begin{axis}[
legend cell align={left},
legend style={fill opacity=0.8, draw opacity=1, text opacity=1, draw=white!80!black},
tick align=outside,
tick pos=left,
title={Dyck},
x grid style={white!69.0196078431373!black},
xmin=-4.15, xmax=109.15,
xtick style={color=black},
xtick={-20,0,20,40,60,80,100,120},
xticklabels={\ensuremath{-}20,0,20,40,60,80,100,120},
y grid style={white!69.0196078431373!black},
ylabel={Difference in Cross Entropy},
ymin=0, ymax=0.404664113482015,
ytick style={color=black},
]
\addplot [semithick, color0]
table {%
1 0.209556805599031
2 0.141897853521801
3 0.117884389498895
4 0.108825735985305
5 0.0848701418980468
6 0.0654647040373133
7 0.0591804544136966
8 0.0797323066155009
9 0.0565933751454614
10 0.0544991234638644
11 0.0538419269298787
12 0.0480385805404995
13 0.0542579988950231
14 0.0459068313629092
15 0.0506115576554459
16 0.0541980289196248
17 0.0384895379121379
18 0.0451154103040587
19 0.0676555555912056
20 0.045449746511522
21 0.044860850664886
22 0.033898779123858
23 0.0356517435348843
24 0.0329012343217058
25 0.0320145534252402
26 0.0309818420049364
27 0.0356263190666727
28 0.0317993291836411
29 0.033956629496955
30 0.0341037609032989
31 0.0395682697424415
32 0.0362916541105552
33 0.0290678249878653
34 0.0296377109264607
35 0.0294654934840217
36 0.0315497570166114
37 0.0332549154458794
38 0.0273797436866897
39 0.029699167655665
40 0.0338140248646996
41 0.026867658725752
42 0.0285503153317906
43 0.0262461971019979
44 0.0235777029019026
45 0.0251048910318168
46 0.0216755241620127
47 0.0205635902826495
48 0.024117746170351
49 0.0246142965102699
50 0.0293920703429333
51 0.0289044058023246
52 0.0340653439674513
53 0.0293435156045708
54 0.027715255114349
55 0.0212629524261417
56 0.0210029279886039
57 0.017910181253985
58 0.0195499870722003
59 0.0201625496919231
60 0.0233571552013632
61 0.0207841678160778
62 0.0179361182610086
63 0.0245820686762043
64 0.0205269432685207
65 0.0223425430523936
66 0.0189782431804109
67 0.0160222543208919
68 0.0212745134041753
69 0.0192978336266875
70 0.0153734031707704
71 0.0177663896640035
72 0.0179034717883713
73 0.0194201724351873
74 0.0228347588202784
75 0.026655094990964
76 0.0203642992490269
77 0.017356679929209
78 0.0145410298695822
79 0.0172384556140546
80 0.0194273181456678
81 0.0189918083417001
82 0.0131692119139784
83 0.0143637677345413
84 0.0147051557522445
85 0.0127953006677031
86 0.0134462117543479
87 0.0224609233788846
88 0.0303821657260153
89 0.0134138127479692
90 0.0181453040324617
91 0.0129068746939577
92 0.0135626431617875
93 0.0122948348833857
94 0.0191362841318707
95 0.0166435780286682
96 0.0189493478267514
97 0.015795777211423
98 0.0134792925155972
99 0.0151143451232068
100 0.0207404220415887
101 0.0161986923835109
102 0.0218978491202271
103 0.0175892312104778
104 0.0224002584438949
};
\addlegendentry{LSTM}
\addplot [semithick, color1]
table {%
1 0.385461889828256
2 0.228195519152578
3 0.0960535700459086
4 0.0572686899566074
5 0.0456568256097994
6 0.0244569282690615
7 0.0141742188780776
8 0.0103861919060707
9 0.0114809848967909
10 0.0124438184667558
11 0.0126462722989437
12 0.00993792331046772
13 0.0132224261638959
14 0.0085191673721271
15 0.0095685446941296
16 0.00485643759575893
17 0.00511519571281771
18 0.00659149673876558
19 0.00462561644584136
20 0.00414100586247657
21 0.004551065296032
22 0.00681568800204957
23 0.00942108966710919
24 0.00505472125612005
25 0.00695274048519945
26 0.00649829822433778
27 0.00405468897733841
28 0.00320571423263938
29 0.0037290076124562
30 0.00361229056276402
31 0.00694748971050874
32 0.004623409966236
33 0.00511491660682928
34 0.00359669047832167
35 0.00688313811839647
36 0.00319174898291696
37 0.00406620446255646
38 0.00488723427987237
39 0.00375853016031957
40 0.00401620817202952
41 0.00337435643902106
42 0.00328126339780088
43 0.00343917022268969
44 0.0040601014444468
45 0.00309488875712627
46 0.00252968321014302
47 0.00743139382238756
48 0.00487113071011747
49 0.00972414951650791
50 0.00314020989796904
51 0.00277413577713004
52 0.00321529736908932
53 0.00341143972432745
54 0.00538463199712946
55 0.0030970798137625
56 0.00431485655476305
57 0.00349347399921962
};
\addlegendentry{Gref}
\addplot [semithick, color2]
table {%
1 0.128078275953707
2 0.0184789197251395
3 0.00867972538920192
4 0.0105067440182063
5 0.00895262641931482
6 0.0253582889606661
7 0.0103699754146301
8 0.00742040560842294
9 0.0123035321764929
10 0.00757729017728226
11 0.00591785287828583
12 0.00535464368988026
13 0.0303000507130176
14 0.0208107937897462
15 0.0104746641406307
16 0.00604238748994934
17 0.0159464006252601
18 0.011101472586201
19 0.00618799587991103
20 0.00410741127284875
21 0.00530932328898692
22 0.00624360626704268
23 0.00425970023121414
24 0.00416938295737967
25 0.00352833076804526
26 0.0115661282548345
27 0.0105105331737915
28 0.00566836883914945
29 0.00373826256910981
30 0.00325086629103588
31 0.0027320896936327
32 0.00435520152017987
33 0.00765053331077059
34 0.00767652811949016
35 0.00519778939264182
36 0.00405196879861591
37 0.00663214157919145
38 0.00308972193748014
39 0.00295351761584306
40 0.00308298744775093
41 0.00410251956806007
42 0.0043877813524994
};
\addlegendentry{JM}
\addplot [semithick, color3]
table {%
1 0.0165244692686763
2 0.00767900159321122
3 0.010359139252179
4 0.00951870638453001
5 0.00500487066272726
6 0.0064751107896488
7 0.00504780859886189
8 0.00815948999692662
9 0.00868123391948394
10 0.00672250909204186
11 0.0061787688102608
12 0.00590048108934893
13 0.00571414855954644
14 0.00342546007271283
15 0.00582997758288939
16 0.00758099276148316
17 0.00606657276138844
18 0.00573685353890918
19 0.00464182505978661
20 0.00384335287443571
21 0.0133088478636272
22 0.00267767126383456
23 0.00434655317088095
24 0.00945603424168662
25 0.00404778114291793
26 0.00381054450270368
27 0.00329079459277504
28 0.00649989866779022
29 0.00569083668530512
30 0.00501433012432217
31 0.00530962522238898
32 0.00396757571319861
33 0.00460047324761548
};
\addlegendentry{NS}
\addplot [semithick, color4]
table {%
1 0.0221405428213096
2 0.00901463012998394
3 0.00730519857573575
4 0.0050258915113286
5 0.00559423607826393
6 0.00393552822311027
7 0.00531209838441005
8 0.00548379445193892
9 0.00451379581403777
10 0.00477382725494224
11 0.0156623315010284
12 0.00487796684439068
13 0.00317675011979357
14 0.00339345581681882
15 0.00286854688945581
16 0.00307866036420323
17 0.00369675881340381
18 0.0037726977500494
19 0.00282917384549786
20 0.00527187048363364
21 0.00380111616851087
22 0.00427130413086274
23 0.00218844245431271
24 0.00409000054633069
25 0.00327814079762467
26 0.00507848698222435
27 0.00338411974724129
28 0.00393584980728057
29 0.007110450747153
30 0.00447798754326967
31 0.00261453135148104
32 0.00397040769802215
33 0.00307135356670596
};
\addlegendentry{NS+S}
\addplot [semithick, color5]
table {%
1 0.210590616029861
2 0.111213545900025
3 0.105614821813008
4 0.0729710683822986
5 0.0111381489662038
6 0.0120400498899319
7 0.00503343700609149
8 0.00657849325053472
9 0.0041164701267874
10 0.00392563980613336
11 0.00658985459661909
12 0.0040754685469726
13 0.00847423301965333
14 0.00709171717622381
15 0.00291036190640914
16 0.003227870396455
17 0.00369932524586591
18 0.00324624330241741
19 0.00306889741493999
20 0.0029505926319634
21 0.00330666877866448
22 0.00240289276513939
23 0.00359864745822658
24 0.00318258330826282
25 0.00160278201923714
26 0.00502703032936336
27 0.00953368082053063
28 0.002934385025398
29 0.00309437781783883
30 0.00141741675308782
31 0.00271300019085197
32 0.00408915828545287
33 0.00259055668319641
34 0.00359011145849808
35 0.00386582161146309
36 0.00399744725493212
37 0.00345941784354675
38 0.00160262228403574
39 0.00157180730103101
40 0.00193821107201753
};
\addlegendentry{NS+U}
\addplot [semithick, color6]
table {%
1 0.0704156448832237
2 0.0107502404290499
3 0.00942779497563673
4 0.00522976723044744
5 0.0114312335283908
6 0.00453090035426762
7 0.00353270160823282
8 0.00458743775202364
9 0.00300298557736967
10 0.00574264175691275
11 0.00340204336250616
12 0.00373321796975101
13 0.00509561687259275
14 0.00951089239511937
15 0.00359742396381302
16 0.00267555142160458
17 0.00673060738551756
18 0.00505653696323627
19 0.00342175024219216
20 0.00281002436459676
21 0.00310746629517566
22 0.00396327032890309
23 0.00266362806045983
24 0.00323259399792708
25 0.00204607004838719
26 0.00333912850777784
27 0.00244682729395662
28 0.0027075974388352
29 0.00163304842309886
30 0.0016404525200735
31 0.0031139343832105
32 0.00243614562957306
33 0.00826704557861491
34 0.00354209754042401
35 0.00286722280586538
36 0.00254109896020815
37 0.00204902093433934
38 0.00310881193603674
39 0.00227990535833111
};
\addlegendentry{NS+S+U}
\legend{}
\end{axis}

\end{tikzpicture}}
    &\scalebox{0.8}{
\begin{tikzpicture}

\definecolor{color0}{rgb}{0.12156862745098,0.466666666666667,0.705882352941177}
\definecolor{color1}{rgb}{1,0.498039215686275,0.0549019607843137}
\definecolor{color2}{rgb}{0.172549019607843,0.627450980392157,0.172549019607843}
\definecolor{color3}{rgb}{0.83921568627451,0.152941176470588,0.156862745098039}
\definecolor{color4}{rgb}{0.580392156862745,0.403921568627451,0.741176470588235}
\definecolor{color5}{rgb}{0.549019607843137,0.337254901960784,0.294117647058824}
\definecolor{color6}{rgb}{0.890196078431372,0.466666666666667,0.76078431372549}

\begin{axis}[
legend cell align={left},
legend style={
  fill opacity=0.8,
  draw opacity=1,
  text opacity=1,
  at={(0.5,0.91)},
  anchor=north,
  draw=white!80!black
},
tick align=outside,
tick pos=left,
title={Dyck},
x grid style={white!69.0196078431373!black},
xmin=37, xmax=103,
xtick style={color=black},
y grid style={white!69.0196078431373!black},
ymin=0, ymax=0.145530964364126,
ytick style={color=black},
]
\addplot [semithick, color0]
table {%
40 0.0924065642509154
42 0.0828644121141485
44 0.0740360885531545
46 0.0679356678658177
48 0.0650548552171427
50 0.0606064949009424
52 0.0576874390640905
54 0.0544102346302596
56 0.0537328528336118
58 0.0521379075747327
60 0.0528063314086739
62 0.0571899543259274
64 0.0494924530103579
66 0.0573067972581321
68 0.0569305548989001
70 0.0580931959927632
72 0.0589204760853353
74 0.0617770235800885
76 0.0635559703718759
78 0.067577164592131
80 0.0661455986310147
82 0.07096370850119
84 0.0811088096013494
86 0.0853039073483189
88 0.0821260305710678
90 0.0855744636550761
92 0.114191599539693
94 0.107400537071778
96 0.0995999720630929
98 0.103208642478841
100 0.140538000108622
};
\addlegendentry{LSTM}
\addplot [semithick, color1]
table {%
40 0.105490060344665
42 0.0908667172558636
44 0.0796187165913489
46 0.0727687691789827
48 0.0647943222930356
50 0.0586659120271679
52 0.054801512619515
54 0.0510832637495777
56 0.0487513989547741
58 0.0462598702998387
60 0.0452093155275263
62 0.0435462136265227
64 0.0440666717603577
66 0.0438071324960052
68 0.0436366378925597
70 0.044888076604559
72 0.0458585513207806
74 0.047316737121755
76 0.0497081872955772
78 0.0520788086783652
80 0.0544918925160454
82 0.0586148992014913
84 0.061643448847673
86 0.0644965034330889
88 0.0682362394896071
90 0.0724897066152411
92 0.075566820053806
94 0.0769731110849353
96 0.081600163348531
98 0.0848742240760628
100 0.0896070266401817
};
\addlegendentry{Gref}
\addplot [semithick, color2]
table {%
40 0.121551178866007
42 0.101691004138131
44 0.0895676098205156
46 0.0792778699103657
48 0.0725562503064284
50 0.0654957220761876
52 0.0594697116908594
54 0.0550163692183278
56 0.0512953014355855
58 0.0477867423204954
60 0.045804858560313
62 0.0446002810868401
64 0.0420900341401654
66 0.0415542611107441
68 0.0406787149985379
70 0.040848546454911
72 0.0409883137351641
74 0.0422477267050884
76 0.0430064954327526
78 0.0443844232946627
80 0.0467749389087924
82 0.049612287190949
84 0.0506709074873789
86 0.0534427139215946
88 0.0554609915085679
90 0.0585079071646918
92 0.0611672484812252
94 0.0627938861671723
96 0.0658059764123197
98 0.0687584274378052
100 0.0708884023363451
};
\addlegendentry{JM}
\addplot [semithick, color3]
table {%
40 0.108612142566464
42 0.0908525230334799
44 0.0801192048725988
46 0.073198664458238
48 0.0673276549620406
50 0.0632324427410159
52 0.057724244414916
54 0.0537001849143505
56 0.050284602079774
58 0.0483082515233556
60 0.0460983075869524
62 0.0445543206453719
64 0.0420524740440116
66 0.0416366130230575
68 0.0407801218146612
70 0.0413341455966364
72 0.0417767207343079
74 0.0436438204550883
76 0.0446424278597007
78 0.0465266182115931
80 0.0494497556526194
82 0.0535538934597745
84 0.0549983431491435
86 0.0587894497333764
88 0.0660672283414331
90 0.068676498339142
92 0.0739213122413059
94 0.0782935777790142
96 0.0835715611417011
98 0.0899395748494213
100 0.0941990459007015
};
\addlegendentry{NS}
\addplot [semithick, color4]
table {%
40 0.0941575884506105
42 0.0781370547376951
44 0.0710468849941266
46 0.0639465658877593
48 0.0598599413140815
50 0.0566572091319474
52 0.0549795049544207
54 0.0514947516757142
56 0.0509713653747302
58 0.0481372703263216
60 0.0477865600203541
62 0.0461374369648162
64 0.0447805389478577
66 0.0447820188066396
68 0.0447915291969075
70 0.0450869652489251
72 0.0456243769843081
74 0.0469477918092551
76 0.0492511687282072
78 0.0505936301775742
80 0.0537322233416627
82 0.0583363435727262
84 0.063286026972673
86 0.0666140164539223
88 0.073583248355478
90 0.0772877942697741
92 0.0832325206586446
94 0.0884374408546722
96 0.0937654641700517
98 0.100014888533133
100 0.105164634158746
};
\addlegendentry{NS+S}
\addplot [semithick, color5]
table {%
40 0.106087133039025
42 0.0930974812456892
44 0.0803804353413489
46 0.0740144577594348
48 0.0667379806157397
50 0.0624199331668738
52 0.0589943745004112
54 0.0547408454257141
56 0.0522537461313092
58 0.0483724729216607
60 0.0471207365060098
62 0.0456783750848561
64 0.0446556891882424
66 0.0423894406816396
68 0.0411009721316901
70 0.0418379693202279
72 0.0407645739021162
74 0.0418845756634216
76 0.0436533095561292
78 0.045243860102416
80 0.0471609825044712
82 0.0520933442316118
84 0.0558126813844377
86 0.0597220669208762
88 0.0663291994615174
90 0.0738866739057631
92 0.0825733934745048
94 0.0901072085355933
96 0.102057033516057
98 0.113555371487679
100 0.126998637871617
};
\addlegendentry{NS+U}
\addplot [semithick, color6]
table {%
40 0.099218444967379
42 0.0802556843706892
44 0.0704924145080156
46 0.0657512845546475
48 0.0618024536705866
50 0.0587049745271682
52 0.0548860405528877
54 0.0524817456387823
56 0.0499600663752785
58 0.0481779466372327
60 0.0465869890674854
62 0.0464906270689829
64 0.0448203526497809
66 0.0451910090118635
68 0.0444530582956394
70 0.0434340300596645
72 0.0448596482813969
74 0.0461295626425884
76 0.0473527692760968
78 0.0500076926775741
80 0.0536979833725268
82 0.0564008672926056
84 0.0599936326712024
86 0.0631127592700143
88 0.0693712465120789
90 0.0736028268934005
92 0.0829682712466824
94 0.0843720625651987
96 0.0940532480779126
98 0.0999668988708859
100 0.106910843935974
};
\addlegendentry{NS+S+U}
\legend{}
\end{axis}

\end{tikzpicture}}
    \\ 
    \scalebox{0.8}{\input{figures/cfl/train/hardest-cfl}} 
    &\scalebox{0.8}{\input{figures/cfl/test/hardest-cfl}}
    \\
    \multicolumn{1}{c}{\begin{minipage}[t]{2.5in}
    \captionof{figure}{Cross-entropy difference in nats between model and source distribution on validation set vs. training time. Each line corresponds to the model which attains the lowest difference in cross-entropy out of all random restarts.}
    \label{fig:cfl-train}
    \end{minipage}}
    &
    \multicolumn{1}{c}{\begin{minipage}[t]{2.5in}
    \captionof{figure}{Cross-entropy difference in nats on the test set, binned by string length. These models are the same as those shown in Figure~\ref{fig:cfl-train}.}
    \label{fig:cfl-test}
    \end{minipage}}
\end{tabular}    
\end{figure*}

\begin{figure} \small
\pgfplotsset{every axis/.style={width=2.8in}}
    \centering
    \begin{tabular}{@{}cc@{}}
\begin{tikzpicture}

\begin{axis}[title={NS+S},axis line style={draw=none},
tick align=outside,
tick pos=left,
xlabel={\(\displaystyle t\)},
xmin=0, xmax=43,
y dir=reverse,
y grid style={white!69.0196078431373!black},
ylabel={$\leftarrow$ Epochs Elapsed},
ymin=0, ymax=30,
ytick style={color=black}, height=5cm
]
\addplot graphics [includegraphics cmd=\pgfimage,xmin=0, xmax=43, ymin=30, ymax=0] {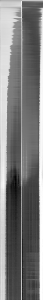};
\end{axis}

\end{tikzpicture} &
\begin{tikzpicture}

\begin{axis}[title={NS+S+U},axis line style={draw=none},
tick align=outside,
tick pos=left,
xlabel={\(\displaystyle t\)},
xmin=0, xmax=43,
y dir=reverse,
y grid style={white!69.0196078431373!black},
ymin=0, ymax=30,
ytick style={color=black},
height=5cm,
]
\addplot graphics [includegraphics cmd=\pgfimage,xmin=0, xmax=43, ymin=30, ymax=0] {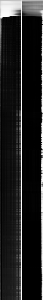};
\end{axis}

\end{tikzpicture}
    \end{tabular}
    \caption{Visualization of the first 30 epochs of training (top to bottom) on the marked reversal task. In each plot, the horizontal axis is the string position (time step). Darkness indicates the weight assigned to the correct stack action type, normalized by the weight of all actions at time~$t$ (black = correct, white = incorrect). The white band in the middle occurs because ``replace'' is considered the correct action type for the middle time step, but the models apparently learned to perform a different action without affecting the results. Both models were trained with learning rate 0.005.}
    \label{fig:weights}
\end{figure}

In Figure \ref{fig:weights}, we show the evolution of stack actions for the NS+S (normalized) and NS+S+U (unnormalized) models over training time on the simplest of the CFL tasks: marked reversal. We see that the normalized model begins solving the task by learning to push and pop symbols close to the middle marker. It then gradually learns to push and pop matching pairs of symbols further and further away from the middle marker. On the other hand, the unnormalized model learns the correct actions for all time steps almost immediately.

\section{Incremental execution}
\label{sec:incremental_execution}

Having demonstrated improvements on synthetic tasks, we now turn to language modeling on natural language. For standard language modeling benchmarks, during both training and evaluation, RNN language models customarily process the entire data set in order as if it were one long sequence, since being able to retain contextual knowledge of past sentences significantly improves predictions for future sentences. Running a full forward and backward pass during training on such a long sequence would be infeasible, so the data set is processed incrementally using a technique called truncated backpropagation through time (BPTT). This technique is feasible for models whose time and space complexity is linear with respect to sequence length, but for memory-augmented models such as stack RNNs, something must be done to limit the time and storage requirements. \Citet{yogatama+al:2018} did this for the superposition stack by limiting the stack to 10 elements. In this section, we propose a technique for limiting the space and time requirements of the \newmodelacronym{} (or NS-RNN), allowing us to use truncated BPTT and retain contextual information.

\subsection{Memory-limited \newmodelacronym{}}
\label{sec:limited_ns}

We introduce the constraint that the stack WFA can only contain transitions $\gamma[i \xrightarrow{} t][q, x \xrightarrow{} r, y]$ where $t - i$ does not exceed a hyperparameter $D$; all other transitions are treated as having zero weight. It may be easier to get an intuition for this constraint in terms of CFGs. The stack WFA formulation is based on Lang's algorithm (\citeyear{lang:1974}), which can be thought of as converting a PDA to a CFG 
and then parsing with a CKY-style algorithm. The equation for $\gamma$ (Equation \ref{eq:gamma}) has three terms, corresponding to rules of the form $(A \rightarrow b)$ (push), $(A \rightarrow Bc)$ (replace), and $(A \rightarrow BCd)$ (pop). The constraint $t-i \leq D$ on $\gamma$ means that these rules can only be used when they span at most $D$ positions.

The equations for $\alpha$ (\ref{eq:alpha1}--\ref{eq:alpha2}) have two cases, which correspond to rules of the form $(A\rightarrow \epsilon)$ and $(A \rightarrow AB)$. The definition of $\alpha$ allows these rules to be used for spans starting at 0 and ending anywhere. This is essentially equivalent to the constraint used in the Hiero machine translation system~\citep{chiang:2005}, which uses synchronous CFGs under the constraint that no nonterminal spans more than 10 symbols, with the exception of so-called glue rules $S \rightarrow X$, $S \rightarrow SX$.

As a consequence of this constraint, if we consider the tensor $\gamma$, which contains the weights of the stack WFA, as a matrix with axes for the variables $i$ and $t$, then the only non-zero entries in $\gamma$ lie in a band of height $D$ along the diagonal. Crucially, column $t$ of $\gamma$ depends only on $\gamma[i \xrightarrow{} t']$ for $t-D \leq i \leq t-2$ and $t-D+1 \leq t' \leq t-1$. Similarly, $\alpha[t]$ depends only on $\alpha[i]$ for $t-D \leq i \leq t-1$ and $\gamma[i \xrightarrow{} t]$ for $t-i \leq D$. So, just as truncated BPTT for an RNN involves freezing the hidden state and forwarding it to the next forward-backward pass, truncated BPTT for the (R)NS-RNN involves forwarding the hidden state of the controller \emph{and} forwarding a slice of $\gamma$ and $\alpha$. This reduces the time complexity of the (R)NS-RNN to $O({|Q|}^4 {|\Gamma|}^3 D^2 n)$ and its space complexity to $O({|Q|}^2 {|\Gamma|}^2 D n)$.

\subsection{Experiments}
\label{sec:ptb_experiments}

Limiting the memory of the (R)NS-RNN now makes it feasbile to run experiments on natural language modeling benchmarks, although the high computational cost of increasing $|Q|$ and $|\Gamma|$ still limits us to settings with little information bandwidth in the stack. We believe this will make it difficult for the (R)NS-RNN to store lexical information on the stack, but it might succeed in using $\Gamma$ as a small set of syntactic categories. To this end, we ran exploratory experiments with the NS-RNN, \newmodelacronym{}, and other language models on the Penn Treebank (PTB) as preprocessed by \citet{mikolov+al:2011}.

We compare four types of model: LSTM, superposition (``JM'') with a maximum stack depth of 10, and memory-limited NS-RNNs (``NS'') and \newmodelacronym{}s (``RNS'') with $D = 35$. We based the hyperparameters for our LSTM baseline and training schedule on those of \citet{semeniuta+al:2016} (details in Appendix~\ref{sec:ptb_hyperparams}).  We also test two variants of JM, pushing either the hidden state or a learned vector. Unless otherwise noted, the LSTM controller has 256 units, one layer, and no dropout. For each model, we randomly search for initial learning rate and gradient clipping threshold; we report results for the model with the best validation perplexity out of 10 random restarts. In addition to perplexity, we also report the recently proposed Syntactic Generalization (SG) score metric \citep{hu+al:2020,gauthier+al:2020}. This score, which ranges from 0 to 1, puts a language model through a battery of psycholinguistically-motivated tests that test how well a model generalizes to non-linear, nested syntactic patterns. \citet{hu+al:2020} noted that perplexity does not, in general, agree with SG score, so we hypothesized the SG score would provide crucial insight into the stack's effectiveness.

\begin{table}
    \caption{Language modeling results on PTB, measured by perplexity and SG score. The setting $|Q| = 1$, $|\Gamma| = 2$ represents minimal capacity in the (R)NS-RNN models and is meant to serve as a baseline for the other settings. The other two settings are meant to test the upper limits of model capacity before computational cost becomes too great. The setting $|Q| = 1$, $|\Gamma| = 11$ represents the greatest number of stack symbol types we can afford to use, using only one PDA state. We selected the setting  $|Q| = 3$, $|\Gamma| = 4$ by increasing the number of PDA states, and then the number of stack symbol types, until computational cost became too great (recall that the time complexity is $O({|Q|}^4 {|\Gamma|}^3)$, so adding states is more expensive than adding stack symbol types).}
    \label{tab:ptb}
    \begin{center} 
    \begin{tabular}{@{}llccc@{}}
        \toprule
        Model & \# Params & Val & Test & SG Score \\
        \midrule
        LSTM, 256 units & 5,656,336 & 125.78 & 120.95 & 0.433 \\
        LSTM, 258 units & 5,704,576 & 122.08 & 118.20 & 0.420 \\
        LSTM, 267 units & 5,922,448 & 125.20 & 120.22 & 0.437 \\
        JM (push hidden state), 247 units & 5,684,828 & \textbf{121.24} & \textbf{115.35} & 0.387 \\
        JM (push learned), $|\mathbf{v}_t| = 22$ & 5,685,289 & 122.87 & 117.93 & 0.431 \\
        NS, $|Q| = 1$, $|\Gamma| = 2$ & 5,660,954 & 126.10 & 122.62 & 0.414 \\
        NS, $|Q| = 1$, $|\Gamma| = 11$ & 5,732,621 & 129.11 & 124.98 & 0.431 \\
        NS, $|Q| = 3$, $|\Gamma| = 4$ & 5,743,700 & 126.71 & 122.53 & 0.447 \\
        RNS, $|Q| = 1$, $|\Gamma| = 2$ & 5,660,954 & 122.64 & 117.56 & 0.435 \\
        RNS, $|Q| = 1$, $|\Gamma| = 11$ & 5,732,621 & 127.21 & 121.84 & 0.386 \\
        RNS, $|Q| = 3$, $|\Gamma| = 4$ & 5,751,892 & 122.67 & 118.09 & 0.408 \\
        \bottomrule
    \end{tabular}
    \end{center}
\end{table}

\paragraph{Results} We show the results of our experiments on the Penn Treebank in Table \ref{tab:ptb}. We reproduce the finding of \citet{yogatama+al:2018} that JM can achieve lower perplexity than an LSTM with a comparable number of parameters, but this does not translate into a better SG score. The results for NS and RNS do not show a clear trend in perplexity or SG score as the number of states or stack symbols increases, or as the modifications in RNS are applied, even when breaking down SG score by type of syntactic test (see Appendix~\ref{sec:full_ptb_results}). We hypothesize that this is due to the information bottleneck caused by using a small discrete set of symbols $\Gamma$ in both models, a limitation we hope to address in future work. The interested reader can find experiments for additional model sizes in Appendix~\ref{sec:full_ptb_results}. For all models we find that SG score is highly variable and uncorrelated to perplexity, corroborating findings by \citet{hu+al:2020}. In fact, when we inspected all randomly searched LSTMs, we found that it is sometimes able to attain scores higher than 0.48 (see Appendix~\ref{sec:full_ptb_results} for details). From this we conclude that improving syntax generalization on natural language remains elusive for all stack RNNs we tested, and that we may need to look beyond cross-entropy/perplexity as a training criterion.

\section{Conclusion}

The \newmodel{} (\newmodelacronym{}) builds upon the strengths of the NS-RNN by letting stack action weights remain unnormalized and providing information about PDA states to the controller. Both of these changes substantially improve learning, allowing the \newmodelacronym{} to surpass other stack RNNs on a range of CFL modeling tasks. Our memory-limited version of the \newmodelacronym{} is a crucial modification towards practical use on natural language. We tested this model on the Penn Treebank, although we did not see performance improvements with the model sizes we were able to test, and in fact no stack RNNs excel in terms of syntactic generalization. We are encouraged by the \newmodelacronym{}'s large improvements on CFL tasks and leave improvements on natural language to future work.

\section*{Reproducibility Statement}

In order to foster reproducibility, we have released all code and scripts used to generate our experimental results and figures at \url{\codeurl}. To ensure that others can replicate our software environment, we developed and ran our code in a Docker container, whose image definition is included in the code repository (see the README for more details). The repository includes the original commands we used to run our experiments, scripts for downloading and preprocessing the PTB dataset of \citet{mikolov+al:2011}, and the test suite definitions needed to compute SG scores. All experimental settings for the CFL experiments are described in \S\ref{sec:cfl_experiments} and our previous paper \citep{dusell+chiang:2020}, and all experimental settings for the PTB experiments may be found in \S\ref{sec:ptb_experiments} and Appendix~\ref{sec:ptb_hyperparams}.

\section*{Acknowledgements}

This research was supported in part by a Google Faculty Research Award to Chiang.

\bibliography{iclr2022_conference}

\begin{thebibliography}{30}
\providecommand{\natexlab}[1]{#1}
\providecommand{\url}[1]{\texttt{#1}}
\expandafter\ifx\csname urlstyle\endcsname\relax
  \providecommand{\doi}[1]{doi: #1}\else
  \providecommand{\doi}{doi: \begingroup \urlstyle{rm}\Url}\fi

\bibitem[Autebert et~al.(1997)Autebert, Berstel, and Boasson]{autebert+:1997}
Jean-Michel Autebert, Jean Berstel, and Luc Boasson.
\newblock Context-free languages and pushdown automata.
\newblock In Grzegorz Rozenberg and Arto Salomaa (eds.), \emph{Handbook of
  Formal Languages}, pp.\  111--174. Springer, 1997.
\newblock \doi{10.1007/978-3-642-59136-5_3}.

\bibitem[Bowman et~al.(2015)Bowman, Manning, and Potts]{bowman+al:2015}
Samuel~R. Bowman, Christopher~D. Manning, and Christopher Potts.
\newblock Tree-structured composition in neural networks without
  tree-structured architectures.
\newblock In \emph{Proc. International Conference on Cognitive Computation
  (CoCo): Integrating Neural and Symbolic Approaches}, pp.\  37--42, 2015.

\bibitem[Bowman et~al.(2016)Bowman, Gauthier, Rastogi, Gupta, Manning, and
  Potts]{bowman+al:2016}
Samuel~R. Bowman, Jon Gauthier, Abhinav Rastogi, Raghav Gupta, Christopher~D.
  Manning, and Christopher Potts.
\newblock A fast unified model for parsing and sentence understanding.
\newblock In \emph{Proceedings of the 54th Annual Meeting of the Association
  for Computational Linguistics (Volume 1: Long Papers)}, pp.\  1466--1477,
  Berlin, Germany, August 2016. Association for Computational Linguistics.
\newblock \doi{10.18653/v1/P16-1139}.
\newblock URL \url{https://aclanthology.org/P16-1139}.

\bibitem[Chiang(2005)]{chiang:2005}
David Chiang.
\newblock A hierarchical phrase-based model for statistical machine
  translation.
\newblock In \emph{Proc.~ACL}, pp.\  263--270, 2005.
\newblock \doi{10.3115/1219840.1219873}.
\newblock URL \url{https://www.aclweb.org/anthology/P05-1033}.

\bibitem[Das et~al.(1992)Das, Giles, and Sun]{das+al:1992}
Sreerupa Das, C.~Lee Giles, and Guo-Zheng Sun.
\newblock Learning context-free grammars: Capabilities and limitations of a
  recurrent neural network with an external stack memory.
\newblock In \emph{Proc. CogSci}, 1992.

\bibitem[DuSell \& Chiang(2020)DuSell and Chiang]{dusell+chiang:2020}
Brian DuSell and David Chiang.
\newblock Learning context-free languages with nondeterministic stack {RNN}s.
\newblock In \emph{Proc. Conference on Computational Natural Language
  Learning}, pp.\  507--519, 2020.
\newblock URL \url{https://www.aclweb.org/anthology/2020.conll-1.41}.

\bibitem[Dyer et~al.(2016)Dyer, Kuncoro, Ballesteros, and Smith]{dyer+al:2016}
Chris Dyer, Adhiguna Kuncoro, Miguel Ballesteros, and Noah~A. Smith.
\newblock Recurrent neural network grammars.
\newblock In \emph{Proc. NAACL HLT}, pp.\  199--209, 2016.
\newblock \doi{10.18653/v1/N16-1024}.
\newblock URL \url{https://aclanthology.org/N16-1024}.

\bibitem[Gauthier et~al.(2020)Gauthier, Hu, Wilcox, Qian, and
  Levy]{gauthier+al:2020}
Jon Gauthier, Jennifer Hu, Ethan Wilcox, Peng Qian, and Roger Levy.
\newblock {S}yntax{G}ym: An online platform for targeted evaluation of language
  models.
\newblock In \emph{Proc. ACL: System Demonstrations}, pp.\  70--76, 2020.
\newblock \doi{10.18653/v1/2020.acl-demos.10}.
\newblock URL \url{https://www.aclweb.org/anthology/2020.acl-demos.10}.

\bibitem[Grefenstette et~al.(2015)Grefenstette, Hermann, Suleyman, and
  Blunsom]{grefenstette+al:2015}
Edward Grefenstette, Karl~Moritz Hermann, Mustafa Suleyman, and Phil Blunsom.
\newblock Learning to transduce with unbounded memory.
\newblock In \emph{Proc. NeurIPS}, volume~2, pp.\  1828--1836, 2015.
\newblock URL
  \url{https://papers.nips.cc/paper/5648-learning-to-transduce-with-unbounded-memory.pdf}.

\bibitem[Greibach(1973)]{greibach:1973}
Sheila~A. Greibach.
\newblock The hardest context-free language.
\newblock \emph{SIAM J. Comput.}, 2\penalty0 (4):\penalty0 304--310, 1973.
\newblock \doi{10.1137/0202025}.

\bibitem[Hao et~al.(2018)Hao, Merrill, Angluin, Frank, Amsel, Benz, and
  Mendelsohn]{hao+al:2018}
Yiding Hao, William Merrill, Dana Angluin, Robert Frank, Noah Amsel, Andrew
  Benz, and Simon Mendelsohn.
\newblock Context-free transductions with neural stacks.
\newblock In \emph{Proc. {B}lackbox{NLP}}, pp.\  306--315, November 2018.
\newblock \doi{10.18653/v1/W18-5433}.
\newblock URL \url{https://www.aclweb.org/anthology/W18-5433}.

\bibitem[Hochreiter \& Schmidhuber(1997)Hochreiter and
  Schmidhuber]{hochreiter+al:1997}
Sepp Hochreiter and J\"{u}rgen Schmidhuber.
\newblock Long short-term memory.
\newblock \emph{Neural Computation}, 9\penalty0 (8):\penalty0 1735--1780,
  November 1997.
\newblock ISSN 0899-7667.
\newblock \doi{10.1162/neco.1997.9.8.1735}.
\newblock URL \url{https://doi.org/10.1162/neco.1997.9.8.1735}.

\bibitem[Hu et~al.(2020)Hu, Gauthier, Qian, Wilcox, and Levy]{hu+al:2020}
Jennifer Hu, Jon Gauthier, Peng Qian, Ethan Wilcox, and Roger Levy.
\newblock A systematic assessment of syntactic generalization in neural
  language models.
\newblock In \emph{Proc. ACL}, pp.\  1725--1744, July 2020.
\newblock URL \url{https://www.aclweb.org/anthology/2020.acl-main.158}.

\bibitem[Joulin \& Mikolov(2015)Joulin and Mikolov]{joulin+mikolov:2015}
Armand Joulin and Tomas Mikolov.
\newblock Inferring algorithmic patterns with stack-augmented recurrent nets.
\newblock In \emph{Proc. NeurIPS}, volume~1, pp.\  190--198, 2015.
\newblock URL
  \url{https://papers.nips.cc/paper/5857-inferring-algorithmic-patterns-with-stack-augmented-recurrent-nets.pdf}.

\bibitem[Kim et~al.(2019)Kim, Dyer, and Rush]{kim+al:2019}
Yoon Kim, Chris Dyer, and Alexander Rush.
\newblock Compound probabilistic context-free grammars for grammar induction.
\newblock In \emph{Proc. ACL}, pp.\  2369--2385, 2019.
\newblock \doi{10.18653/v1/P19-1228}.
\newblock URL \url{https://aclanthology.org/P19-1228}.

\bibitem[Lafferty et~al.(2001)Lafferty, McCallum, and Pereira]{lafferty+:2001}
John~D. Lafferty, Andrew McCallum, and Fernando C.~N. Pereira.
\newblock Conditional random fields: Probabilistic models for segmenting and
  labeling sequence data.
\newblock In \emph{Proceedings of the Eighteenth International Conference on
  Machine Learning}, pp.\  282–--289, 2001.

\bibitem[Lang(1974)]{lang:1974}
Bernard Lang.
\newblock Deterministic techniques for efficient non-deterministic parsers.
\newblock In \emph{Proc.~Colloquium on Automata, Languages, and Programming},
  pp.\  255--269, 1974.
\newblock \doi{10.1007/978-3-662-21545-6_18}.

\bibitem[McCoy et~al.(2020)McCoy, Frank, and Linzen]{mccoy+al:2020}
Richard McCoy, Robert~H. Frank, and Tal Linzen.
\newblock Does syntax need to grow on trees? {S}ources of hierarchical
  inductive bias in sequence-to-sequence networks.
\newblock \emph{Trans. ACL}, 8:\penalty0 125--140, 2020.
\newblock URL
  \url{https://www.mitpressjournals.org/doi/pdf/10.1162/tacl_a_00304}.

\bibitem[Merity et~al.(2018)Merity, Keskar, and Socher]{merity+al:2018}
Stephen Merity, Nitish~Shirish Keskar, and Richard Socher.
\newblock Regularizing and optimizing {LSTM} language models.
\newblock In \emph{Proc. ICLR}, 2018.
\newblock URL \url{https://openreview.net/forum?id=SyyGPP0TZ}.

\bibitem[Merrill et~al.(2019)Merrill, Khazan, Amsel, Hao, Mendelsohn, and
  Frank]{merrill+al:2019}
William Merrill, Lenny Khazan, Noah Amsel, Yiding Hao, Simon Mendelsohn, and
  Robert Frank.
\newblock Finding hierarchical structure in neural stacks using unsupervised
  parsing.
\newblock In \emph{Proceedings of the 2019 ACL Workshop BlackboxNLP: Analyzing
  and Interpreting Neural Networks for NLP}, pp.\  224--232, Florence, Italy,
  August 2019. Association for Computational Linguistics.
\newblock URL \url{https://www.aclweb.org/anthology/W19-4823}.

\bibitem[Mikolov et~al.(2011)Mikolov, Deoras, Kombrink, Burget, and
  Cernock{\'y}]{mikolov+al:2011}
Tomas Mikolov, Anoop Deoras, Stefan Kombrink, L.~Burget, and J.~Cernock{\'y}.
\newblock Empirical evaluation and combination of advanced language modeling
  techniques.
\newblock In \emph{Proc. INTERSPEECH}, 2011.

\bibitem[Nangia \& Bowman(2018)Nangia and Bowman]{nangia+bowman:2018}
Nikita Nangia and Samuel Bowman.
\newblock {L}ist{O}ps: A diagnostic dataset for latent tree learning.
\newblock In \emph{Proc.~NAACL~Student Research Workshop}, pp.\  92--99, 2018.
\newblock \doi{10.18653/v1/N18-4013}.
\newblock URL \url{https://www.aclweb.org/anthology/N18-4013}.

\bibitem[Semeniuta et~al.(2016)Semeniuta, Severyn, and
  Barth]{semeniuta+al:2016}
Stanislau Semeniuta, Aliaksei Severyn, and Erhardt Barth.
\newblock Recurrent dropout without memory loss.
\newblock In \emph{Proc. COLING}, pp.\  1757--1766, 2016.
\newblock URL \url{https://www.aclweb.org/anthology/C16-1165}.

\bibitem[Shen et~al.(2019{\natexlab{a}})Shen, Tan, Hosseini, Lin, Sordoni, and
  Courville]{shen+al:2019}
Yikang Shen, Shawn Tan, Arian Hosseini, Zhouhan Lin, Alessandro Sordoni, and
  Aaron~C Courville.
\newblock Ordered memory.
\newblock In \emph{Proc. NeurIPS}, volume~32, 2019{\natexlab{a}}.
\newblock URL
  \url{https://proceedings.neurips.cc/paper/2019/file/d8e1344e27a5b08cdfd5d027d9b8d6de-Paper.pdf}.

\bibitem[Shen et~al.(2019{\natexlab{b}})Shen, Tan, Sordoni, and
  Courville]{shen+al:2018}
Yikang Shen, Shawn Tan, Alessandro Sordoni, and Aaron Courville.
\newblock Ordered neurons: Integrating tree structures into recurrent neural
  networks.
\newblock In \emph{Proc. ICLR}, 2019{\natexlab{b}}.
\newblock URL \url{https://openreview.net/forum?id=B1l6qiR5F7}.

\bibitem[Sun et~al.(1995)Sun, Giles, Chen, and Lee]{sun+al:1995}
G.~Z. Sun, C.~Lee Giles, H.~H. Chen, and Y.~C. Lee.
\newblock The neural network pushdown automaton: Model, stack, and learning
  simulations.
\newblock Technical Report UMIACS-TR-93-77 and CS-TR-3118, University of
  Maryland, 1995.
\newblock URL \url{https://arxiv.org/abs/1711.05738}.
\newblock revised version.

\bibitem[Suzgun et~al.(2019)Suzgun, Gehrmann, Belinkov, and
  Shieber]{suzgun+:2019}
Mirac Suzgun, Sebastian Gehrmann, Yonatan Belinkov, and Stuart~M. Shieber.
\newblock Memory-augmented recurrent neural networks can learn generalized
  {D}yck languages, 2019.
\newblock URL \url{https://arxiv.org/abs/1911.03329}.
\newblock {a}rXiv:1922.03329.

\bibitem[van Schijndel et~al.(2019)van Schijndel, Mueller, and
  Linzen]{schijndel+al:2019}
Marten van Schijndel, Aaron Mueller, and Tal Linzen.
\newblock Quantity doesn{'}t buy quality syntax with neural language models.
\newblock In \emph{Proc. EMNLP-IJCNLP}, pp.\  5831--5837, 2019.
\newblock \doi{10.18653/v1/D19-1592}.
\newblock URL \url{https://www.aclweb.org/anthology/D19-1592}.

\bibitem[Wilcox et~al.(2019)Wilcox, Levy, and Futrell]{wilcox+al:2019}
Ethan Wilcox, Roger Levy, and Richard Futrell.
\newblock Hierarchical representation in neural language models: Suppression
  and recovery of expectations.
\newblock In \emph{Proc. BlackboxNLP}, pp.\  181--190, August 2019.
\newblock \doi{10.18653/v1/W19-4819}.
\newblock URL \url{https://www.aclweb.org/anthology/W19-4819}.

\bibitem[Yogatama et~al.(2018)Yogatama, Miao, Melis, Ling, Kuncoro, Dyer, and
  Blunsom]{yogatama+al:2018}
Dani Yogatama, Yishu Miao, G{\'a}bor Melis, Wang Ling, Adhiguna Kuncoro, Chris
  Dyer, and Phil Blunsom.
\newblock Memory architectures in recurrent neural network language models.
\newblock In \emph{Proc. ICLR}, 2018.
\newblock URL \url{https://openreview.net/pdf?id=SkFqf0lAZ}.

\end{thebibliography}
\bibliographystyle{iclr2022_conference}

\appendix

\section{Baseline stack RNNs}

\subsection{Stratification stack}
\label{sec:stratification_details}

We implement the stratification stack of \citet{grefenstette+al:2015} with the following equations. In the original definition, the controller produces a hidden state $\mathbf{h}_t$ and a separate output $\mathbf{o}'_t$ that is used to compute $a_t$ and $\mathbf{y}_t$, but for simplicity and parity with \citet{dusell+chiang:2020}, we set $\mathbf{o}'_t = \mathbf{h}_t$. Let $m = |\mathbf{v}_t|$ be the stack embedding size.
\begin{align*}
    a_t = \textsc{Actions}(\mathbf{h}_t) &= (u_t, d_t, \mathbf{v}_t) \\
    u_t &= \sigma(\Affine{hu}{\mathbf{h}_t}) \\
    d_t &= \sigma(\Affine{hd}{\mathbf{h}_t}) \\
    \mathbf{v}_t &= \tanh(\Affine{hv}{\mathbf{h}_t}) \\
    s_t = \textsc{Stack}(s_{t-1}, a_t) &= (V_t, \mathbf{s}_t) \\
    V_t[i] &= \begin{cases}
        V_{t-1}[i] & 1 \leq i < t \\
        \mathbf{v}_t & i = t
    \end{cases} \\
    \mathbf{s}_t[i] &= \begin{cases}
        \max(0, \mathbf{s}_{t-1}[i] - \max(0, u_t - \sum_{j=i+1}^{t-1} \mathbf{s}_{t-1}[j])) & 1 \leq i < t \\
        d_t & i = t
    \end{cases} \\
    V_0 &\text{ is a } 0 \times m \text{ matrix} \\
    \mathbf{s}_0 &\text{ is a vector of size 0} \\
    \mathbf{r}_t = \textsc{Reading}(s_t) &= \sum_{i=1}^t (\min(\mathbf{s}_t[i], \max(0, 1 - \sum_{j=i+1}^t \mathbf{s}_t[j]))) \cdot V_t[i]
\end{align*}

\citet{yogatama+al:2018} noted that the strafication stack can implement multiple pops per time step by allowing $u_t > 1$, although the push action immediately following would still be conditioned on the previous stack top $\mathbf{r}_t$. \Citet{hao+al:2018} augmented this model with differentiable queues that allow it to buffer input and output and act as a transducer. \Citet{merrill+al:2019} experimented with variations of this model where $u_t = 1$ and $d_t \in (0, 4)$, $d_t = 1$ and $u_t \in (0, 4)$, and $u_t \in (0, 4)$ and $d_t \in (0, 1)$.

\subsection{Superposition stack}
\label{sec:superposition_details}

We implement the superposition stack of \citet{joulin+mikolov:2015} with the following equations. We deviate slightly from the original definition by adding the bias terms $\bparam{ha}$ and $\bparam{hv}$. The original definition also connects the controller to multiple stacks that push \emph{scalars}; instead, we push a vector to a single stack, which is equivalent to multiple scalar stacks whose push/pop actions are synchronized. The original definition includes the top $k$ stack elements in the stack reading, but we only include the top element. We also treat the value of the bottom of the stack as 0 instead of $-1$.
\begin{align*}
    a_t = \textsc{Actions}(\mathbf{h}_t) &= (\mathbf{a}_t, \mathbf{v}_t) \\
    \mathbf{a}_t &= \begin{bmatrix}
        a^{\mathrm{push}}_t \\
        a^{\mathrm{noop}}_t \\
        a^{\mathrm{pop}}_t
    \end{bmatrix} = \mathrm{softmax}(\Affine{ha}{\mathbf{h}_t}) \\
    \mathbf{v}_t &= \sigma(\Affine{hv}{\mathbf{h}_t}) \\
    \textsc{Stack}(s_{t-1}, a_t) &= s_t \\
    s_t[i] &= \begin{cases}
        \mathbf{v}_{t+1} & i = 0 \\
        a^{\mathrm{push}}_t s_{t-1}[i-1] + a^{\mathrm{noop}}_t s_{t-1}[i] + a^{\mathrm{pop}}_t s_{t-1}[i+1] & 0 < i \leq t \\
        \mathbf{0} & i > t
    \end{cases} \\
    \mathbf{r}_t = \textsc{Reading}(s_t) &= s_t[1]
\end{align*}

\Citet{yogatama+al:2018} developed an extension to this model called the Multipop Adaptive Computation Stack that executes a variable number of pops per time step, up to a fixed limit $K$. They also restricted the stack to a maximum size of 10 elements, where the bottom element of a full stack is discarded when a new element is pushed; in other words, $s_t[i] = \mathbf{0}$ for $i > K$. \Citet{suzgun+:2019} experimented with a modification of the parameterization of $\mathbf{a}_t$ and different softmax operators for normalizing the weights used to compute $\mathbf{a}_t$.

\section{Details of formal language experiments}
\label{sec:formal_hyperparams}

For every training run, we sample a training set of 10,000 strings from the PCFG, with lengths drawn uniformly from $[40, 80]$. Similarly, we sample a validation set of 1,000 strings with lengths drawn uniformly from $[40, 80]$. For each task, we sample a test set of 100 strings per length for each length in $[40, 100]$. Whereas the training and validation sets are randomized for each experiment, the test sets are the same across all models and random restarts.

In all cases, the LSTM has a single layer with 20 hidden units. We grid-search the initial learning rate from $\{0.01, 0.005, 0.001, 0.0005\}$. For Gref and JM, we search for stack vector element sizes in $\{2, 20, 40\}$ (the pushed vector in JM is learned). For the NS models, we manually choose a small number of PDA states and stack symbol types based on how we would expect a PDA to solve the task. For marked reversal, unmarked reversal, and Dyck, we use $|Q| = 2$ and $|\Gamma| = 3$; and for padded reversal and hardest CFL, we use $|Q| = 3$ and $|\Gamma| = 3$. For each hyperparameter setting searched, we run five random restarts. For each type of model, we select the model with the lowest difference in cross-entropy between the model and true distribution on the validation set. We use the same initialization and optimization settings as in our earlier paper \citep{dusell+chiang:2020} and train for a maximum of 200 epochs.

\section{Wall-clock training time}
\label{sec:wall_clock_time}

We report wall-clock execution time for each model on the marked reversal task in Table \ref{tab:wall_clock_time}. We ran the LSTM, Gref, and JM models in CPU mode, as this was faster than running on GPU due to the small model size. We ran experiments for the NS models in GPU mode on a pool of the following NVIDIA GPU models, automatically selected based on availability: GeForce GTX TITAN X, TITAN X (Pascal), and GeForce GTX 1080 Ti.

\begin{table}
    \caption{Wall-clock execution time for each model on the marked reversal task, measured in seconds per epoch of training (averaged over all epochs). The speed of the NS models is roughly the same; there is some variation here due to differences in training data and GPU model.}
    \label{tab:wall_clock_time}
    \begin{center}
    \begin{tabular}{@{}lc@{}}
        \toprule
        Model & Time per epoch (s) \\
        \midrule
        LSTM & 51 \\
        Gref & 801 \\
        JM & 169 \\
        NS & 1022 \\
        NS+S & 980 \\
        NS+U & 960 \\
        NS+S+U & 1060 \\
        \bottomrule
    \end{tabular}
    \end{center}
\end{table}

\section{Details of natural language experiments}
\label{sec:ptb_hyperparams}

The hyperparameters for our baseline LSTM, initialization, and optimization scheme are based on the unregularized LSTM experiments of \citet{semeniuta+al:2016}. We train all models using simple stochastic gradient descent (SGD) as recommended by prior language modeling work \citep{merity+al:2018} and truncated BPTT with a sequence length of 35. For all models, we use a minibatch size of 32. We randomly initialize all parameters uniformly from the interval $[-0.05, 0.05]$. We divide the learning rate by 1.5 whenever the validation perplexity does not improve, and we stop training after 2 epochs of no improvement in validation perplexity.

For each model, we randomly search for initial learning rate and gradient clipping threshold; we report results for the model with the best validation perplexity out of 10 randomly searched models. The learning rate, which is divided by batch size and sequence length, is drawn from a log-uniform distribution over $[1, 100]$, and the gradient clipping threshold, which is multiplied by batch size and sequence length, is drawn from a log-uniform distribution over $[1 \times 10^{-5}, 1 \times 10^{-3}]$. (We scale the learning rate and gradient clipping threshold this way because, under our implementation, sequence length and batch size can vary when the data set is not evenly divisible by the prescribed values. Other language modeling papers follow a different scaling convention for these two hyperparameters, typically scaling the learning rate by sequence length but not by batch size, and not rescaling the gradient clipping threshold. Under this convention the learning rate would be drawn from $[0.03125, 3.125]$ and the gradient clipping threshold from $[0.0112, 1.12]$.)

\section{Additional results for natural language experiments}
\label{sec:full_ptb_results}

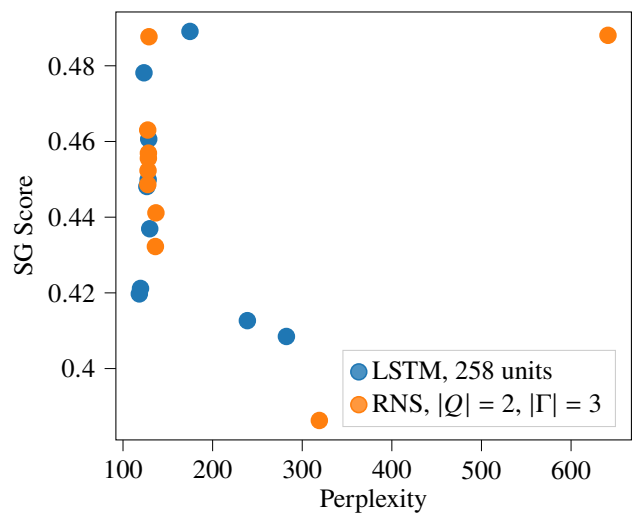
\begin{figure}
    \centering
\begin{tikzpicture}

\definecolor{color0}{rgb}{0.12156862745098,0.466666666666667,0.705882352941177}
\definecolor{color1}{rgb}{1,0.498039215686275,0.0549019607843137}

\begin{axis}[
legend cell align={left},
legend style={
  fill opacity=0.8,
  draw opacity=1,
  text opacity=1,
  at={(0.97,0.03)},
  anchor=south east,
  draw=white!80!black
},
tick align=outside,
tick pos=left,
x grid style={white!69.0196078431373!black},
xlabel={Perplexity},
xmin=92.0524, xmax=667.3436,
xtick style={color=black},
y grid style={white!69.0196078431373!black},
ylabel={SG Score},
ymin=0.381148801406238, ymax=0.494218855769199,
ytick style={color=black}
]
\addplot [semithick, color0, mark=*, mark size=3, mark options={solid}, only marks]
table {%
238.776 0.412661443353663
129.812 0.436939048323488
123.39 0.4781457514009
119.519 0.421174807902497
282.34 0.408479050612918
174.717 0.48907930784361
128.046 0.449913871195336
128.695 0.460627592807227
118.202 0.419725449988608
126.619 0.448103171301112
};
\addlegendentry{LSTM, 258 units}
\addplot [semithick, color1, mark=*, mark size=3, mark options={solid}, only marks]
table {%
319.083 0.386288349331828
136.137 0.432247326154649
127.572 0.463023913538788
127.655 0.448621815298017
127.915 0.452302665975435
136.72 0.441142794403664
128.344 0.456925994168557
128.958 0.487675105638492
128.308 0.455557784447945
641.194 0.488048183700358
};
\addlegendentry{RNS, $|Q| = 2$, $|\Gamma| = 3$}
\end{axis}

\end{tikzpicture}
    \caption{SG score vs. test perplexity, shown on all 10 random restarts for an LSTM and an RNS-RNN. SG score is uncorrelated with perplexity, and models that narrowly miss out on having the best perplexity often have much higher SG scores.}
    \label{fig:sg_score_vs_perplexity}
\end{figure}

In Table \ref{tab:full_ptb} we show additional experimental results on the Penn Treebank. In Table \ref{tab:circuits} we show the same experiments with SG score broken down by syntactic ``circuit'' as defined by \citet{hu+al:2020}, offering a more fine-grained look at the classes of errors the models make. We see that SG score is highly variable and does not follow the same trends as perplexity. In Figure \ref{fig:sg_score_vs_perplexity}, we plot SG score vs. test perplexity for all 10 random restarts of an LSTM and an RNS-RNN. We see that many of the models that were not selected actually have a much higher SG score (even above 0.48), suggesting that the standard validation perplexity criterion is a poor choice for syntactic generalization.

\begin{table}
    \caption{Language modeling results on PTB, measured by perplexity and SG score, with additional experiments included.}
    \label{tab:full_ptb}
    \begin{center}
    \begin{tabular}{@{}llccc@{}}
        \toprule
        Model & \# Params & Val & Test & SG Score \\
        \midrule
        LSTM, 256 units & 5,656,336 & 125.78 & 120.95 & 0.433 \\
        LSTM, 258 units & 5,704,576 & 122.08 & 118.20 & 0.420 \\
        LSTM, 267 units & 5,922,448 & 125.20 & 120.22 & 0.437 \\
        JM (push hidden state), 247 units & 5,684,828 & \textbf{121.24} & \textbf{115.35} & 0.387 \\
        JM (push learned), $|\mathbf{v}_t| = 22$ & 5,685,289 & 122.87 & 117.93 & 0.431 \\
        NS, $|Q| = 1$, $|\Gamma| = 2$ & 5,660,954 & 126.10 & 122.62 & 0.414 \\
        NS, $|Q| = 1$, $|\Gamma| = 3$ & 5,664,805 & 123.41 & 119.25 & 0.430 \\
        NS, $|Q| = 1$, $|\Gamma| = 4$ & 5,669,684 & 121.66 & 117.91 & 0.432 \\
        NS, $|Q| = 1$, $|\Gamma| = 5$ & 5,675,591 & 123.01 & 119.54 & 0.452 \\
        NS, $|Q| = 1$, $|\Gamma| = 6$ & 5,682,526 & 129.94 & 125.45 & 0.432 \\
        NS, $|Q| = 1$, $|\Gamma| = 7$ & 5,690,489 & 126.11 & 121.94 & 0.443 \\
        NS, $|Q| = 1$, $|\Gamma| = 11$ & 5,732,621 & 129.11 & 124.98 & 0.431 \\
        NS, $|Q| = 2$, $|\Gamma| = 2$ & 5,668,664 & 128.16 & 123.52 & 0.412 \\
        NS, $|Q| = 2$, $|\Gamma| = 3$ & 5,680,996 & 129.51 & 126.00 & \textbf{0.471} \\
        NS, $|Q| = 2$, $|\Gamma| = 4$ & 5,697,440 & 124.28 & 120.18 & 0.433 \\
        NS, $|Q| = 2$, $|\Gamma| = 5$ & 5,717,996 & 124.24 & 119.34 & 0.429 \\
        NS, $|Q| = 3$, $|\Gamma| = 2$ & 5,681,514 & 125.32 & 120.62 & 0.470 \\
        NS, $|Q| = 3$, $|\Gamma| = 3$ & 5,707,981 & 122.96 & 118.89 & 0.420 \\
        NS, $|Q| = 3$, $|\Gamma| = 4$ & 5,743,700 & 126.71 & 122.53 & 0.447 \\
        RNS, $|Q| = 1$, $|\Gamma| = 2$ & 5,660,954 & 122.64 & 117.56 & 0.435 \\
        RNS, $|Q| = 1$, $|\Gamma| = 3$ & 5,664,805 & 121.83 & 116.46 & 0.430 \\
        RNS, $|Q| = 1$, $|\Gamma| = 4$ & 5,669,684 & 127.99 & 123.06 & 0.437 \\
        RNS, $|Q| = 1$, $|\Gamma| = 5$ & 5,675,591 & 126.41 & 122.25 & 0.441 \\
        RNS, $|Q| = 1$, $|\Gamma| = 6$ & 5,682,526 & 122.57 & 117.79 & 0.416 \\
        RNS, $|Q| = 1$, $|\Gamma| = 7$ & 5,690,489 & 123.51 & 120.48 & 0.430 \\
        RNS, $|Q| = 1$, $|\Gamma| = 11$ & 5,732,621 & 127.21 & 121.84 & 0.386 \\
        RNS, $|Q| = 2$, $|\Gamma| = 2$ & 5,670,712 & 122.11 & 117.22 & 0.399 \\
        RNS, $|Q| = 2$, $|\Gamma| = 3$ & 5,684,068 & 131.46 & 127.57 & 0.463 \\
        RNS, $|Q| = 2$, $|\Gamma| = 4$ & 5,701,536 & 124.96 & 121.61 & 0.431 \\
        RNS, $|Q| = 2$, $|\Gamma| = 5$ & 5,723,116 & 122.92 & 117.87 & 0.423 \\
        RNS, $|Q| = 3$, $|\Gamma| = 2$ & 5,685,610 & 129.48 & 124.66 & 0.433 \\
        RNS, $|Q| = 3$, $|\Gamma| = 3$ & 5,714,125 & 127.57 & 123.00 & 0.434 \\
        RNS, $|Q| = 3$, $|\Gamma| = 4$ & 5,751,892 & 122.67 & 118.09 & 0.408 \\
        \bottomrule
    \end{tabular}
    \end{center}
\end{table}

\begin{table}
    \caption{SG scores broken down by circuit. Agr. = Agreement, Lic. = Licensing, GPE = Garden-Path Effects, GSE = Gross Syntactic Expectation, CE = Center Embedding, LDD = Long-Distance Dependencies.}
    \label{tab:circuits}
    \begin{center}
    \begin{tabular}{@{}lcccccc@{}}
        \toprule
        Model & Agr. & Lic. & GPE & GSE & CE & LDD \\
        \midrule
        LSTM, 256 units & 0.667 & 0.446 & 0.330 & 0.397 & 0.482 & 0.414 \\
        LSTM, 258 units & 0.658 & 0.447 & 0.335 & 0.375 & 0.518 & 0.357 \\
        LSTM, 267 units & 0.667 & 0.497 & 0.343 & 0.446 & 0.411 & 0.350 \\
        JM (push hidden state) & 0.640 & 0.408 & 0.296 & 0.310 & 0.464 & 0.352 \\
        JM (push learned) & 0.684 & 0.439 & 0.340 & 0.408 & 0.482 & 0.395 \\
        NS, $|Q| = 1$, $|\Gamma| = 2$ & 0.588 & 0.452 & 0.298 & 0.391 & 0.339 & 0.418 \\
        NS, $|Q| = 1$, $|\Gamma| = 3$ & 0.623 & 0.467 & 0.400 & 0.413 & 0.393 & 0.354 \\
        NS, $|Q| = 1$, $|\Gamma| = 4$ & 0.640 & 0.497 & 0.331 & 0.375 & 0.571 & 0.340 \\
        NS, $|Q| = 1$, $|\Gamma| = 5$ & 0.605 & 0.514 & 0.394 & 0.413 & 0.589 & 0.344 \\
        NS, $|Q| = 1$, $|\Gamma| = 6$ & 0.632 & 0.424 & 0.408 & 0.391 & 0.464 & 0.399 \\
        NS, $|Q| = 1$, $|\Gamma| = 7$ & 0.719 & 0.470 & 0.351 & 0.473 & 0.500 & 0.344 \\
        NS, $|Q| = 1$, $|\Gamma| = 11$ & 0.640 & 0.432 & 0.329 & 0.424 & 0.500 & 0.413 \\
        NS, $|Q| = 2$, $|\Gamma| = 2$ & 0.702 & 0.388 & 0.329 & 0.446 & 0.446 & 0.371 \\
        NS, $|Q| = 2$, $|\Gamma| = 3$ & 0.658 & 0.527 & 0.367 & 0.446 & 0.518 & 0.411 \\
        NS, $|Q| = 2$, $|\Gamma| = 4$ & 0.632 & 0.464 & 0.345 & 0.386 & 0.518 & 0.387 \\
        NS, $|Q| = 2$, $|\Gamma| = 5$ & 0.711 & 0.464 & 0.307 & 0.413 & 0.518 & 0.355 \\
        NS, $|Q| = 3$, $|\Gamma| = 2$ & 0.711 & 0.528 & 0.349 & 0.435 & 0.518 & 0.406 \\
        NS, $|Q| = 3$, $|\Gamma| = 3$ & 0.746 & 0.439 & 0.316 & 0.375 & 0.411 & 0.376 \\
        NS, $|Q| = 3$, $|\Gamma| = 4$ & 0.702 & 0.450 & 0.364 & 0.484 & 0.536 & 0.369 \\
        RNS, $|Q| = 1$, $|\Gamma| = 2$ & 0.702 & 0.460 & 0.280 & 0.451 & 0.464 & 0.404 \\
        RNS, $|Q| = 1$, $|\Gamma| = 3$ & 0.649 & 0.427 & 0.438 & 0.418 & 0.446 & 0.347 \\
        RNS, $|Q| = 1$, $|\Gamma| = 4$ & 0.658 & 0.412 & 0.342 & 0.565 & 0.339 & 0.418 \\
        RNS, $|Q| = 1$, $|\Gamma| = 5$ & 0.728 & 0.449 & 0.370 & 0.429 & 0.482 & 0.371 \\
        RNS, $|Q| = 1$, $|\Gamma| = 6$ & 0.614 & 0.422 & 0.314 & 0.435 & 0.518 & 0.377 \\
        RNS, $|Q| = 1$, $|\Gamma| = 7$ & 0.649 & 0.460 & 0.374 & 0.337 & 0.411 & 0.404 \\
        RNS, $|Q| = 1$, $|\Gamma| = 11$ & 0.614 & 0.447 & 0.291 & 0.266 & 0.446 & 0.338 \\
        RNS, $|Q| = 2$, $|\Gamma| = 2$ & 0.649 & 0.417 & 0.365 & 0.375 & 0.339 & 0.334 \\
        RNS, $|Q| = 2$, $|\Gamma| = 3$ & 0.640 & 0.474 & 0.411 & 0.446 & 0.554 & 0.408 \\
        RNS, $|Q| = 2$, $|\Gamma| = 4$ & 0.658 & 0.469 & 0.336 & 0.326 & 0.500 & 0.403 \\
        RNS, $|Q| = 2$, $|\Gamma| = 5$ & 0.693 & 0.420 & 0.339 & 0.370 & 0.607 & 0.376 \\
        RNS, $|Q| = 3$, $|\Gamma| = 2$ & 0.579 & 0.435 & 0.295 & 0.440 & 0.554 & 0.445 \\
        RNS, $|Q| = 3$, $|\Gamma| = 3$ & 0.632 & 0.444 & 0.356 & 0.418 & 0.482 & 0.403 \\
        RNS, $|Q| = 3$, $|\Gamma| = 4$ & 0.588 & 0.427 & 0.342 & 0.353 & 0.482 & 0.373 \\
        \bottomrule
    \end{tabular}
    \end{center}
\end{table}

\end{document}